\documentclass{article} % For LaTeX2e
\usepackage{paper_conference,times}

\usepackage{amsmath,amsfonts,bm}
\usepackage{multirow}
\usepackage{tabularx}
\usepackage{graphicx}
\usepackage{adjustbox}
\usepackage{amsthm}
\usepackage{xspace}

\newcommand{\mset}[1]{\left\{\kern-.5em\left\{ #1 \right\}\kern-.5em\right\}}
\newcommand{\mmset}[1]{\{\kern-.4em\{ #1 \}\kern-.4em\}}

\newcommand{\parr}[1]{\left (#1\right )}

\def\eqref#1{equation~\ref{#1}}
\def\1{\bm{1}}

\def\vc{{\bm{c}}}
\def\vd{{\bm{d}}}

\def\vo{{\bm{o}}}
\def\vp{{\bm{p}}}

\def\vr{{\bm{r}}}

\def\vw{{\bm{w}}}
\def\vx{{\bm{x}}}

\def\vz{{\bm{z}}}

\DeclareMathAlphabet{\mathsfit}{\encodingdefault}{\sfdefault}{m}{sl}
\SetMathAlphabet{\mathsfit}{bold}{\encodingdefault}{\sfdefault}{bx}{n}

\usepackage{hyperref}
\usepackage{url}
\usepackage{bm}
\usepackage{wrapfig}
\usepackage{graphicx}
\usepackage{cleveref}
\usepackage{wrapfig}
\usepackage{tabularx}
\usepackage{adjustbox}
\usepackage{siunitx}
\usepackage{amsmath}
\usepackage{multicol}
\usepackage{color}
\usepackage{booktabs}
\usepackage{subcaption,amsfonts,dcolumn}
\title{StyleNeRF: A Style-based 3D-Aware Generator for High-resolution Image Synthesis}
\iclrfinalcopy

\def \mpi{$^\ddag$}
\def \nus{$^\diamond$}
\def \fair{$^\dagger$}

\author{
Jiatao Gu\fair ,
Lingjie Liu\mpi\thanks{corresponding author.}  , 
Peng Wang\nus, 
Christian Theobalt\mpi
\\
\fair Facebook AI \ \ \
\mpi Max Planck Institute for Informatics \ \ \
\nus The University of Hong Kong \\
\fair\texttt{jgu@fb.com}  \ \ \
\mpi\texttt{\{lliu,theobalt\}@mpi-inf.mpg.de}\ \ \
\nus\texttt{pwang3@cs.hku.hk}\\
}

\begin{document}

\maketitle

\begin{abstract}
We propose \emph{StyleNeRF}, a 3D-aware generative model for photo-realistic high-resolution image synthesis with high multi-view  consistency, which can be trained on unstructured 2D images. 
Existing approaches either cannot synthesize high-resolution images with fine details or yield noticeable 3D-inconsistent artifacts. In addition, many of them lack control over style attributes and explicit 3D camera poses.
\emph{StyleNeRF} integrates the neural radiance field (NeRF) into a style-based generator to tackle the aforementioned challenges, i.e., improving rendering efficiency and 3D consistency for high-resolution image generation.
We perform volume rendering only to produce a low-resolution feature map and progressively apply upsampling in 2D to address the first issue. 
To mitigate the inconsistencies caused by 2D upsampling, we propose multiple designs, including a better upsampler and a new regularization loss. With these designs, \emph{StyleNeRF} can synthesize high-resolution images at interactive rates while preserving 3D consistency at high quality. 
\emph{StyleNeRF} also enables control of camera poses and different levels of styles, which can generalize to unseen views. It also supports challenging tasks, including zoom-in and-out, style mixing, inversion, and semantic editing.\footnote{Please check our video at:
\url{http://jiataogu.me/style_nerf/}.}
\end{abstract}

\section{Introduction}
\begin{figure}[h]
    \centering
    \includegraphics[width=\linewidth]{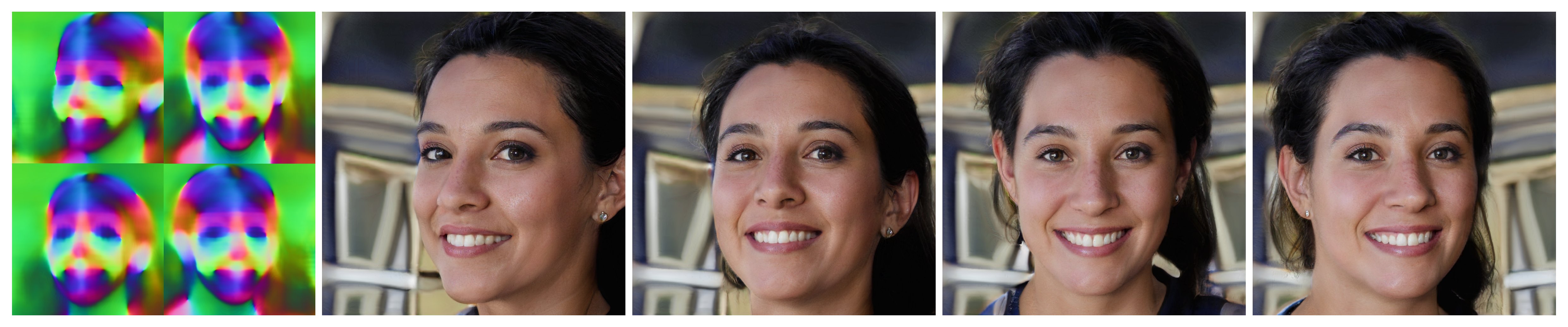}
    \caption{Synthesized $1024^2$ images by \emph{StyleNeRF}, with the corresponding low-resolution feature maps. \emph{StyleNeRF} can generate photo-realistic high-resolution images from novel views at interactive rates while preserving high 3D consistency. None of existing methods can achieve both features. }
    \label{fig:pipeline}
    %\vspace{-7pt}
\end{figure}

Photo-realistic free-view image synthesis of real-world scenes is a long-standing problem in computer vision and computer graphics. Traditional graphics pipeline requires production-quality 3D models, computationally expensive rendering, and manual work, making it challenging to apply to large-scale image synthesis for a wide range of real-world scenes. 
In the meantime, Generative Adversarial Networks~\citep[GANs,][]{gan} can be trained on a large number of unstructured images to synthesize high-quality images.
However, most GAN models operate in 2D space. Therefore, they lack the 3D understanding of the training images, which results in their inability to synthesize images of the same 3D scene with multi-view consistency.
They also lack direct 3D camera control over the generated images.

Natural images are the 2D projection of the 3D world. Hence, recent works on generative models~\citep{schwarz2020graf,chan2021pi} enforce 3D structures by incorporating a neural radiance field~\citep[NeRF,][]{mildenhall2020nerf}. 
However, these methods cannot synthesize high-resolution images with delicate details due to the computationally expensive rendering process of NeRF. 
Furthermore, the slow rendering process leads to inefficient training and makes these models unsuitable for interactive applications. 
GIRAFFE~\citep{niemeyer2021giraffe} combines NeRF with a CNN-based renderer, which has the potential to synthesize high-resolution images. 
However, this method falls short of 3D-consistent image generation and so far has not shown high-resolution results.

We propose \emph{StyleNeRF}, a new 3D-aware generative model for high-resolution 3D consistent image synthesis at interactive rates. % preserving high 3D consistency,
It also allows control of the 3D camera pose and enables control of specific style attributes. 
\emph{StyleNeRF} incorporates 3D scene representations into a style-based generative model. To prevent the expensive direct color image rendering from the original NeRF approach, we only use NeRF to produce a low-resolution feature map and upsample it progressively to high resolution.
To improve 3D consistency, we propose several designs, including a desirable upsampler that achieves high consistency while mitigating artifacts in the outputs, a novel regularization term that forces the output to match the rendering result of the original NeRF and fixing the issues of view direction condition and noise injection. \emph{StyleNeRF} is trained using unstructured real-world images. A progressive training strategy significantly improves the stability of learning real geometry. 

We evaluate \emph{StyleNeRF} on various challenging datasets.  \emph{StyleNeRF} can synthesize photo-realistic $1024^2$ images at interactive rates while achieving high multi-view consistency. None of the existing methods can achieve both characteristics. Additionally, \emph{StyleNeRF} enables direct control on styles, and 3D camera poses even for the poses starkly different from training and supports applications including style mixing, interpolation, inversion, and semantic editing.

\section{Related Work}
\vspace{-8pt}\paragraph{Neural Implicit Fields}
Representing 3D scenes as neural implicit fields has increasingly gained much attention.  
\citet{Michalkiewicz_2019_ICCV,OccupancyNetworks,park2019deepsdf,Peng2020ConvolutionalON} predict neural implicit fields with 3D supervision. %\citet{lombardi2019neural,mildenhall2020nerf,liu2020neural,zhang2020nerf++} have shown that one can learn neural implicit fields from 2D images only. 
Some of them~\citep{sitzmann2019scene,niemeyer2019differentiable} assume that the ray color only lies on the geometry surface and propose differentiable renderers to learn a neural implicit surface representation. 
NeRF and its variants~\citep{mildenhall2020nerf,liu2020neural,zhang2020nerf++} utilize a volume rendering technique to render neural implicit volume representations for novel view synthesis. 
% To improve the surface quality extracted from the learned scene representations, some recent works apply volume rendering to learning implicit surface representations. 
In this work, we propose a generative variant of NeRF~\citep{mildenhall2020nerf}. Unlike the discussed methods, which require posed multi-view images, our approach only needs unstructured single-view images for training. 

% NeRF NV  NSVF etc

% no generalization

\vspace{-8pt}\paragraph{Image Synthesis with GANs}
% Past years have witnessed great progress in 2D generative models. 
% Several GAN models~\citep{Goodfellow14,GMAN,Mordido2018DropoutGANLF,Doan2019OnlineAC,sagan,Brock2018Biggan,karras2018progressive} can synthesize high-resolution images, but they have limited controls over synthesized results.
Starting from~\citet{gan}, GANs have demonstrated  high-quality  results~\citep{GMAN,Mordido2018DropoutGANLF,Doan2019OnlineAC,sagan,Brock2018Biggan,karras2018progressive}.  StyleGANs~\citep{karras2019style,karras2020analyzing} achieve SOTA quality and support different levels of style control. 
\citet{karras2021alias} solve the ``texture sticking" problem of 2D GANs in generating animations with 2D transformations. 
Some methods~\citep{harkonen2020ganspace, tewari2020stylerig, shen2020interpreting,abdal2020image2stylegan++, tewari2020pie,FreeStyleGAN2021,shoshan2021gancontrol}  leverage disentangled properties in the latent space to enable explicit controls, most of which focus on faces. While these methods can synthesize face poses parameterized by two angles, extending them to general objects and controlling 3D cameras is not easy. \cite{chen2021sofgan} proposed to generate segmentation maps from implicit fields to enable 3D control. However, it requires 3D meshes for pre-training.
In contrast, our work can synthesize images for \emph{general objects}, enabling \emph{explicit 3D camera} control.%\LJ{Jiatao, please check, cite SofGAN}. 

\vspace{-8pt}\paragraph{3D-Aware GANs}
Recently, neural scene representations have been integrated into 2D generative models to enable direct camera control. Voxel-based  GANs~\citep{henzler2019platonicgan,nguyen2019hologan,nguyen2020blockgan} 
lack fine details in the results due to the voxel resolution restriction.
Radiance fields-based methods~\citep{schwarz2020graf,chan2021pi} have higher quality and better 3D consistency but have difficulties training for high-resolution images ($512^2$ and beyond) due to the expensive rendering process of radiance fields. GIRAFFE~\citep{niemeyer2021giraffe} %, a compositional generative model, 
improved the training and rendering efficiency by combining NeRF with a ConvNet-based renderer.
%for a low-resolution feature map rendering and upsampling the feature map with $3 \times 3$ convolution layers. 
% While GIRAFFE enabled scene composition, 
However, it produced severe view-inconsistent artifacts due to its particular network designs (e.g. $3 \times 3$ \texttt{Conv}, upsampler). In contrast, our method can effectively preserve view consistency in image synthesis.

\section{Method}
%In this section, we introduce \textit{\emph{StyleNeRF}} -- an efficient 3D-aware generator with style control and can be trained only using unstructured 2D images with adversarial objectives. 
\subsection{Image synthesis as neural implicit field rendering} 
\paragraph{Style-based Generative Neural Radiance Field}
We start by modeling a 3D scene as neural radiance field~\citep[NeRF,][]{mildenhall2020nerf}. It is typically parameterized as multilayer perceptrons (MLPs),  % multilayer perceptron (MLP)
which takes the position $\bm{x} \in \mathbb{R}^3$ and viewing direction $\bm{d} \in \mathbb{S}^2$ as input, and predicts the density $\sigma(\bm{x}) \in \mathbb{R}^+$ and view-dependent color $\bm{c}(\bm{x},\bm{d}) \in \mathbb{R}^3$. 
To model high-frequency details, follwing NeRF~\citep{mildenhall2020nerf}, we map each dimension of $\bm{x}$ and $\bm{d}$ with Fourier features : 
% To better modeling high-frequency details, \citet{mildenhall2020nerf} proposed to map each dimension of $\bm{x}$ and $\bm{d}$ with Fourier features:
\begin{equation}
    \zeta^L(x) = \left[\sin(2^0x), \cos(2^0x), \ldots, \sin(2^{L-1}x), \cos(2^{L-1}x)\right]
    \label{eq.pos}
\end{equation}
%Without loss of generality\LJ{what's generality?}, 
We formalize \emph{StyleNeRF} representations % for any spatial point $\bm{x}$ 
by conditioning NeRF with style vectors $\bm{w}$ as follows:
\begin{equation}
    \bm{\phi}_{\vw}^n(\bm{x}) = g^{n}_{\vw}\circ g^{n-1}_{\vw}\circ \ldots \circ g^{1}_{\vw} \circ \zeta\left(\bm{x}\right), \ \ \text{where} \ \  \bm{w} = f(\bm{z}), \vz \in \mathcal{Z}
\end{equation}
Similar as StyleGAN2~\citep{karras2020analyzing}, $f$ is a mapping network that maps noise vectors from the spherical Gaussian space $\mathcal{Z}$ to the style space $\mathcal{W}$;
$g^i_{\vw}(.)$ is the $i^\text{th}$ layer MLP 
%which consists of an affine transformation followed by the nonlinear activation.
whose weight matrix is modulated by the input style vector $\bm{w}$. % \LJ{moved to beginning: similar as in StyleGAN2~\citep{karras2020analyzing}}.
$\bm{\phi}_{\vw}^n(\bm{x})$ is the $n$-th layer feature of that point. We then use the extracted features to predict the density and color, respectively:
\begin{equation}
    \sigma_{\vw}(\bm{x}) = h_{\sigma}\circ\bm{\phi}^{n_\sigma}_{\vw}(\bm{x}), \ \ \  \bm{c}_\vw(\bm{x},\bm{d}) = h_{c}\circ\left[\bm{\phi}^{n_c}_\vw(\bm{x}), \zeta\left(\bm{d}\right)\right],
    \label{eq:sigma_color}
\end{equation}
where $h_{\sigma}$ and $h_c$ can be a linear projection or 2-layer MLPs. % \LJ{-> can be a linear function or encoded by 2-layer MLPs}. 
Different from the original NeRF,
we assume $n_c > n_\sigma$ for \Cref{eq:sigma_color} % which makes sense 
as the visual appearance generally needs more capacity to model than the geometry. %\LJ{we can move this sentence upfront to Eq.3}
%\LJ{not only in our formulation, NeRF also has this property, so I suggest removing `in our formulation'}
% $n_\sigma$ and $n_c$ do not have to be the same, and 
The first $\min(n_\sigma, n_c)$ layers are shared in the network.% for $\sigma$ and $\bm{c}$. % $\sigma$ and $\bm{c}$ 
%\LJ{-> ..., and the first few ($\min(n_\sigma, n_c)$) layers are shared in the networks for predicting $\sigma$ and $\bm{c}$. }.
\vspace{-5pt}\paragraph{Volume Rendering}
Image synthesis is modeled as volume rendering from a given camera pose $\vp\in\mathcal{P}$. 
For simplicity, we assume a camera is located on the unit sphere pointing to the origin with a fixed field of view (FOV). We sample the camera's pitch \& yaw from a uniform or Gaussian distribution depending on %the camera distribution approximation of 
the dataset.  
To render an image $I\in \mathbb{R}^{H\times W\times 3}$, %we first locate the corresponding camera ray $\bm{r}(t)=\bm{o} + t\bm{d}$ ($\vo$ is the camera origin) of each pixel, and calculate the final output using volume rendering equation 
we shoot a camera ray $\bm{r}(t)=\bm{o} + t\bm{d}$ ($\vo$ is the camera origin) for each pixel, and then calculate the color using the volume rendering equation:
\begin{equation}
    I^{\text{NeRF}}_\vw(\bm{r}) = \int_0^\infty\!\! p_\vw(t)\vc_\vw(\vr(t),\vd) dt, \ \  \text{where} \ \  p_\vw(t) = \exp\parr{ - \int_0^t\!\!  \sigma_\vw(\vr(s)) ds }\cdot\sigma_\vw(\vr(t))
    \label{eq:rendering}
\end{equation}
In practice, the above equation is discretized by accumulating sampled points along the ray. Following NeRF~\citep{mildenhall2020nerf}, stratified and hierarchical sampling are applied for more accurate discrete approximation to the continuous implicit function. 
% Compared to 2D generative models~\citep{karras2019style,karras2020analyzing}, this process generates pixels independently, without employing any up-sampling or information exchange with neighboring pixels (e.g. convolutions), which leads to 3D consistency in the output.
% this process does not involve any up-sampling or information exchange with neighboring pixels (e.g. convolutions), and all pixels are generated independently. \LJ{..., this volume rendeirng process generates pixels independently, without employing any up-sampling or information exchange with neighboring pixels (e.g. convolutions), which leads to multi-view consistency in the synthesized results. }

\vspace{-5pt}\paragraph{Challenges} 
Compared to 2D generative models (e.g., StyleGANs~\citep{karras2019style,karras2020analyzing}), the images generated by NeRF-based models have 3D consistency, which is guaranteed by modeling the image synthesis as a physics process, and the neural 3D scene representation is invariant across different viewpoints. 
However, the drawbacks are apparent: these models cost much more computation to render an image at the exact resolution. 
For example, 2D GANs are $100\sim1000$ times more efficient to generate a $1024^2$ image than NeRF-based models.
Furthermore, NeRF consumes much more memory to cache the intermediate results for gradient back-propagation during training, making it difficult to train on high-resolution images.
Both of these restrict the scope of applying NeRF-based models in high-quality image synthesis, especially at the training stage when calculating the objective function over the whole image is crucial. % It also makes training inefficient. 

\subsection{Approximation for high-resolution image  generation}
In this section, we propose how to improve the efficiency of \emph{StyleNeRF} by taking inspiration from 2D GANs. %in the meanwhile, preserving the inherent 3D consistency to the furthest extent. 
We observe that the image generation of 2D GANs (e.g., StyleGANs) is fast due to two main reasons: (1) each pixel only takes single forward pass through the network; (2) image features are generated progressively from coarse to fine, and the feature maps with higher resolutions typically have a smaller number of channels to save memory. 

In \emph{StyleNeRF}, the first point can be partially achieved by early aggregating the features into the 2D space before the final colors are computed. 
%\LJ{add: After the early aggregation, 
In this way, each pixel is assigned with a feature vector, % which will be used for further processing,
%each pixel 
Furthermore, it only needs to pass through a network once rather than calling the network multiple times for all sampled points on the ray as NeRF does. % }
% Using early aggregation to 
We approximate \Cref{eq:rendering} as:
\begin{equation}
    I^{\text{Approx}}_\vw(\bm{r}) = \int_0^\infty\!\!  p_\vw(t)\cdot 
    %\vc_\vw(\vr(t),\vd)
    h_{c}\circ\left[\bm{\phi}_\vw^{n_c}(\vr(t)), \zeta\left(\bm{d}\right)\right]
    dt \approx h_{c}\circ\left[ 
    \bm{\phi}^{n_c,n_\sigma}_\vw
    \left(
        \mathcal{A}(\vr)% \int_0^\infty\!\!  p_\vw(t)\cdot \bm{\phi}_{\vw}^{n_\sigma}(\vr(t)) dt
    \right), 
    \zeta\left(\bm{d}\right)
    \right],
    \label{eq:agg}
\end{equation}
where $\bm{\phi}^{n,n_\sigma}_\vw (\mathcal{A}(\vr)) = g^{n}_{\vw}\circ g^{n-1}_{\vw}\circ \ldots \circ g^{n_\sigma+1}_{\vw} \circ \mathcal{A}(\vr)$ and
$\mathcal{A}(\vr)=\int_0^\infty\!\!  p_\vw(t)\cdot \bm{\phi}_{\vw}^{n_\sigma}(\vr(t)) dt$.
% Here, we assume $n_c > n_\sigma$ for \Cref{eq:sigma_color} which makes sense as the visual appearance generally needs more capacity to model than the geometry. \LJ{we can move this sentence upfront to Eq.3}
% We denote  $\bm{\phi}^{n,n_\sigma}_\vw(\mathcal{A}(\vr))$ as the features at the $n^\text{th}$ layer with $\mathcal{A}(\vr)$ passing through $(n-n_\sigma)$ MLP layers. 
The definitions of $\mathcal{A}(.)$ and $\bm{\phi}^{n,n_\sigma}_\vw(.)$ can be extended to the operations on a set of rays, each ray processed independently. 
% Moreover, the early aggregation strategy enables us to leverage techniques (described in point (2)) from 2D GANs for further rendering acceleration. 
Next, instead of using volume rendering to render a high-resolution feature map directly, we can employ NeRF to generate a downsampled feature map at a low resolution and then employ upsampling in 2D space to progressively increase into the required high resolution. 
We take two adjacent resolutions as an example.  
Suppose $R_L\in \mathbb{R}^{H/2\times W/2}$ and $R_H\in \mathbb{R}^{H\times W}$ are the corresponding rays of the pixels in the low- and high-resolution images, respectively. 
% For each ray $\vr \in R$, let $\mathcal{A}(\vr)$ be the aggregated features of $\vr$ produced by NeRF, i.e., $\mathcal{A}(\vr)=\int_0^\infty\!\!  p_\vw(t)\cdot \bm{\phi}_{\vw}^{n_\sigma}(\vr(t)) dt$. 
% For convenience, we use $\bm{\phi}^{n,n_\sigma}_\vw(\vr)$ to represent the $n^\text{th}$ layer aggregated 2D feature for each ray $\vr$.
To approximate the high-resolution feature map, we can up-sample in the low-resolution feature space:
\begin{equation}
    \bm{\phi}^{n,n_\sigma}_\vw(\mathcal{A}(R_H))\approx \texttt{Upsample}\left(\bm{\phi}^{n,n_\sigma}_\vw(\mathcal{A}(R_L))\right)
    \label{eq:up}
\end{equation}
Recursively inserting \texttt{Upsample} operators enables efficient high-resolution image synthesis as the computationally expensive volume rendering only needs to generate a low-resolution feature map. 
% For instance, for rendering an $H\times W$ image with $K$ sampled points on each ray, if we upsample features every two layers, the effective number of layers is reduced from $n_cKHW$ to $\left(1.67+{\left(n_\sigma K - 1.67\right)}/{(n_c-n_\sigma)^2}\right)HW$, and we only need few rays to start with.  
%\LJ{For instance, ... is not that clear.}
% $(1.66 + 0.03K)HW$ 
% ($K$ is the number of samples to approximate the integral in \Cref{eq:agg}). 
% since our input resolution is only 
The efficiency is further improved when using fewer channels for higher resolution.
%\LJ{have we done this in our network?}

While early aggregation and upsampling operations can accelerate the rendering process for high-resolution image synthesis, they would come with scarification to the inherent consistency of NeRF.
There are two reasons why they introduce inconsistency. First, the resulting model contains nonlinear transformations
to capture spurious correlations in 2D observation, mainly when substantial ambiguity exists. For example, our training data are unstructured single-view images without sufficient multi-view supervision. 
Second, such a pixel-space operation like up-sampling would compromise 3D consistency. 
Therefore, na\"ive model designs would lead to severe multi-view inconsistent outputs (e.g., when moving the camera to render images, hairs are constantly changing). 
In the following, we propose several designs and choices to alleviate the inconsistency in the outputs. 

% \vspace{-5pt}\paragraph{Choose a right upsampler} 
\subsection{Preserving 3D consistency} 
\begin{figure}[t]
    \centering
   \centering
    \includegraphics[width=\linewidth]{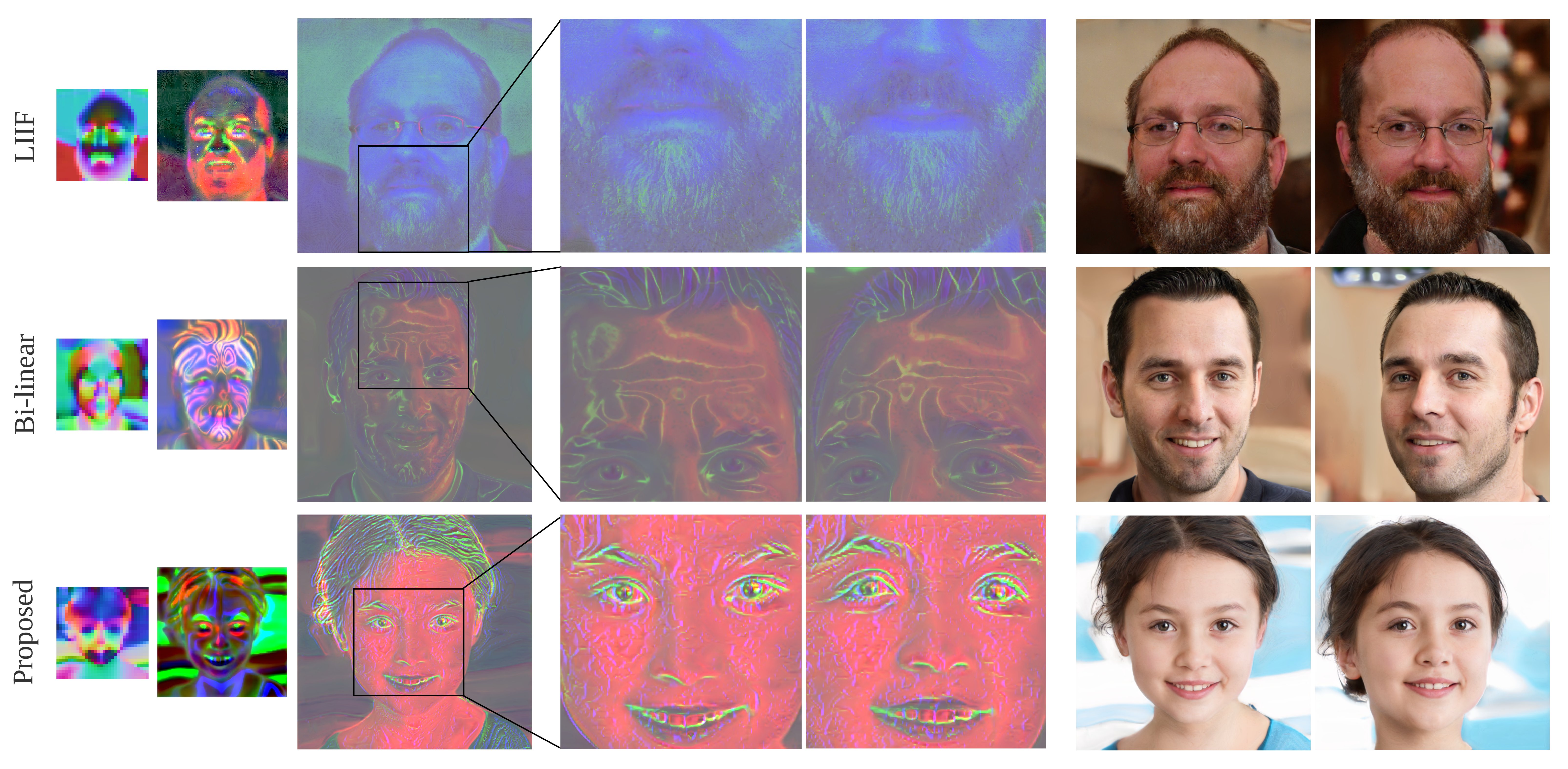}
    \caption{Internal representations and the outputs from \emph{StyleNeRF} trained with different upsampling operators. With LIIF, patterns stick to the same pixel coordinates when the viewpoint changes; With bilinear interpolation, bubble-shape artifacts can be seen on the feature maps and images. Our proposed upsampler faithfully preserves 3D consistency while getting rid of bubble-shape artifacts. }
    \label{fig:upsample}
    \vspace{-12pt}
\end{figure}
\vspace{-5pt}\paragraph{Upsampler design} 
Up-sampling in 2D space causes multi-view inconsistency in general; however, the specific design choice of the upsampler determines how much such inconsistency is introduced. 
% The first principle to tackle this is smoothness.
% We use MLP ($1\times1$ Conv) as the basic building block of \emph{StyleNeRF}, given that pixel-independent processing is better for preserving multi-view consistency. 
As our model is directly derived from NeRF, MLP ($1\times1$ \texttt{Conv}) is the basic building block.
With MLPs, however, pixel-wise learnable upsamplers such as pixelshuffle~\citep{shi2016real} or LIIF~\citep{chen2021learning} produce  ``chessboard'' or ``texture sticking'' artifacts due to its tendency of relying on the image coordinates implicitly. %\LJ{it has this artifact due to MLPs?}
% As the basic building block of \emph{StyleNeRF} is MLP ($1\times1$ Conv), pixel-wise learnable upsampler like pixelshuffle~\citep{shi2016real} or LIIF~\citep{chen2021learning} produces  ``chessboard'' or ``sticking-texture'' artifacts due to its tendency of relying on the image coordinates implicitly. 
In the meanwhile, \citet{karras2019style,karras2020analyzing,karras2021alias} proposed to use non-learnable upsamplers that  interpolate the feature map with pre-defined low-pass filters (e.g. bilinear interpolation). % which interpolates the feature maps into high resolution. 
While these upsamplers can produce smoother outputs, we observed non-removable ``bubble'' artifacts
% throughout the training phase \LJ{remove ``throughout the training phase"} 
in both the feature maps and output images. We conjecture it is due to the lack of local variations when combining MLPs with fixed low-pass filters. We achieve the balance between consistency and image quality by combining these two approaches (see \Cref{fig:upsample}). For any input feature map $X\in \mathbb{R}^{N\times N\times D}$:
\begin{equation}
    \texttt{Upsample}(X) = \texttt{Conv2d}\left({\texttt{Pixelshuffle}}\left(\texttt{Repeat}(X, 4) + \psi_\theta(X), 2\right), K\right),
\end{equation}
where $\psi_\theta: \mathbb{R}^D \rightarrow \mathbb{R}^{4D}$ is a learnable 2-layer MLP, and $K$ is a fixed  blur kernel~\citep{zhang2019making}.%  Please see the visual comparison in~\Cref{fig:upsample}.

\vspace{-5pt}\paragraph{NeRF path regularization}
We propose a new regularization term to enforce 3D consistency, which regularizes the model output to match the original path (\Cref{eq:rendering}). In this way, the final outputs can be closer to the NeRF results, which have multi-view consistency. 
% We sub-sample pixels on the output images and compare against those generated by NeRF. 
This is implemented by sub-sampling pixels on the output and comparing them against those generated by NeRF:
% In addition to the architecture change, we also propose to enforce the 3D consistency by regularizing the model output to match the original path (\Cref{eq:rendering}) without early aggregation and upsampling. This is implemented by sub-sampling pixels on the output:
\begin{equation}
    \mathcal{L}_{\text{NeRF-path}} = \frac{1}{|S|}\sum\nolimits_{(i,j)\in S} 
    \left(
    I_{\vw}^{\text{Approx}}(R_\text{in})[i, j] - I_{\vw}^{\text{NeRF}}(R_\text{out}[i, j])
    \right)^2,
    \label{eq:nerf_path}
\end{equation}
where $S$ is the set of randomly sampled pixels; $R_\text{in}$ and $R_\text{out}$ are the corresponding rays of the pixels in the low-resolution image generated via NeRF and high-resolution output of \emph{\emph{StyleNeRF}}.
\vspace{-5pt}\paragraph{Remove view direction condition} %The most straightforward way to enforce consistency is to discard the condition on view direction which was 
Predicting colors with view direction condition was suggested by the original NeRF for modeling view-dependent effects and was by default applied in most follow-up works~\citep{chan2021pi,niemeyer2021giraffe}. 
However, this design would give the model additional freedom to capture spurious correlations and dataset bias, especially if only a  single-view target is provided. We, therefore, remove the view direction in color prediction, which improves the synthesis consistency (See \Cref{fig:ablation}). 
\vspace{-5pt}\paragraph{Fix 2D noise injection} 
Existing studies~\citep{karras2019style,karras2020analyzing,feng2020noise} have showed that injecting per-pixel noise %after each convolution 
can increase the model's capability of modeling stochastic variation (e.g. hairs, stubble). Nevertheless, such 2D noise only exists in the image plane, which will not change in a 3D consistent way when the camera moves. To preserve 3D consistency, our default solution is to trade the model's capability of capturing variation by removing the noise injection as \citet{karras2021alias} does. 
Optionally, we also propose a novel geometry-aware noise injection based on the estimated surface from \emph{\emph{StyleNeRF}}. See \Cref{sec.noise} for more details.
% \LJ{\emph{StyleNeRF} or NeRF}. See the  Appendix for more details. 
% \JG{TODO: spherical noise? }

\subsection{StyleNeRF Architecture}
\begin{figure}[t]
    \centering
   \centering
    \includegraphics[width=\linewidth]{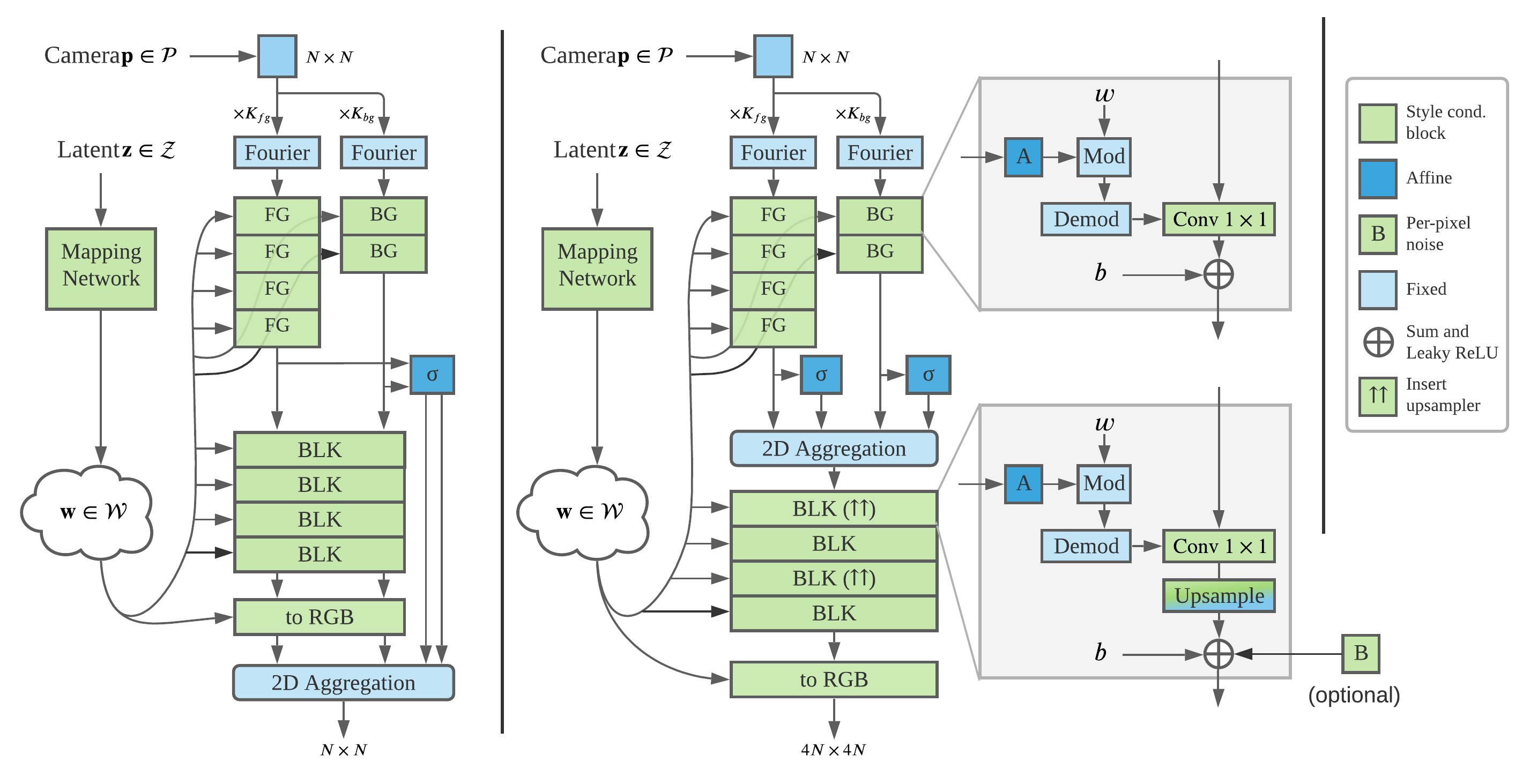}
    \caption{\label{fig:arch}Example architecture. (Left) Original NeRF path; (right) Main path of \emph{\emph{StyleNeRF}}.}
    \vspace{-10pt}
\end{figure}
In this section, we describe the network architecture and the learning procedure of \emph{StyleNeRF}. % An overview of the \emph{StyleNeRF} generator is shown in \Cref{fig:arch}.
\vspace{-5pt}\paragraph{Mapping Network}
Following StyleGAN2, latent codes are sampled from standard Gaussian and processed by a mapping network. Finally, the output vectors are broadcast to the synthesis networks. 

\vspace{-5pt}\paragraph{Synthesis Network}
Considering that our training images generally have unbounded background, we choose NeRF++~\citep{zhang2020nerf++}, a variant of NeRF, as the \emph{StyleNeRF} backbone. NeRF++ consists of a foreground NeRF in a unit sphere and a background NeRF represented with inverted sphere parameterization. 
% Points in the background NeRF are sampled based on inverse depth and calculated using inverse sphere parameterization.
As shown in \Cref{fig:arch}, two MLPs are used to predict the density where the background network has fewer parameters than the foreground one. Then a shared MLP is employed for color prediction. % Different from the conventional StyleGAN
% By default, no skip connections are used.
Each style-conditioned block consists of an affine transformation layer and a $1\times 1$ convolution layer (\texttt{Conv}).
The \texttt{Conv} weights are modulated with the affine-transformed styles, and then demodulated for computation. %\LJ{what is transformed styles?}
\texttt{leaky\_ReLU} is used as non-linear activation. 
% Different from \citet{chan2021pi}, our early experiments showed that \texttt{leaky\_ReLU} instead of \texttt{SIREN}~\citep{sitzmann2020implicit} is more stable as the activation to our model, and achieves better qualitative results \LJ{suggest removing this sentence.}.
The number of blocks depends on the input and target image resolutions.

\vspace{-5pt}\paragraph{Discriminator \& Objectives}
We use the same discriminator as StyleGAN2. % The discriminator is based on $3\times 3$ convolution and residual networks. 
% As \emph{StyleNeRF} is able to generate the whole image efficiently, there is no need to adopt a sophisticated patch-based discriminator~\citep{schwarz2020graf}. 
Following previous works~\citep{chan2021pi,niemeyer2021giraffe}, \emph{\emph{StyleNeRF}} adopts a non-saturating GAN objective with R1 regularization~\citep{mescheder2018training}. A new NeRF path regularization is employed to enforce 3D consistency. 
The final loss function is defined as follows ($D$ is the discriminator and $G$ is the generator including the mapping and synthesis networks):
\begin{equation}
    \mathcal{L}(D, G)=\mathbb{E}_{\vz\sim\mathcal{Z},\vp\sim\mathcal{P}}\left[f(D(G(\vz,\vp))\right] + \mathbb{E}_{I\sim p_{\text{data}}}\left[f(-D(I) + \lambda \|\nabla D(I)\|^2)\right] + \beta\cdot \mathcal{L}_{\text{NeRF-path}}
\end{equation}
where $f(u)=-\log(1+\exp(-u))$, and $p_{\text{data}}$ is the data distribution. We set $\beta=0.2$ and $\lambda=0.5$.
\vspace{-5pt}\paragraph{Progressive Training}
We train \emph{\emph{StyleNeRF}} progressively from low to high resolution, which makes the training more stable and efficient. 
We observed in the experiments that were directly training for the highest resolution easily 
%causes model drifting \LJ{is it well-defined?} and thus 
makes the model fail to capture the object geometry. 
We suspect it is because both \Cref{eq:agg,eq:up} are just approximations to the original NeRF. % \LJ{are you sure this is the reason? If just a guess, suggest removing this sentence.}
% As both \Cref{eq:agg,eq:up} are just approximations to the original NeRF, directly training with the highest resolution may cause model drifting, which fails to capture the object geometry.
Therefore, inspired by ~\citet{karras2017progressive}, we propose a new \textit{three}-stage progressive training  strategy:
%Let the input and target resolutions as $R_{\text{in}}$ and $R_{\text{out}}$, respectively. 
For the first $T_1$ images, we train \emph{StyleNeRF} without approximation at low-resolution; then, %new layers are faded linearly into 
during $T_1\sim T_2$ images,
both the generator and discriminator linearly increase the output resolutions until reaching the target resolution; At last, we fix the architecture and continue training the model at the highest resolution until $T_3$ images. Please refer to the \Cref{sec.progressive} for more details.

%\begin{wrapfigure}{l}{0.75\textwidth}

%\end{wrapfigure}
    
% Different from StyleGAN2

\begin{figure}[t]
    \centering
    \includegraphics[width=\linewidth]{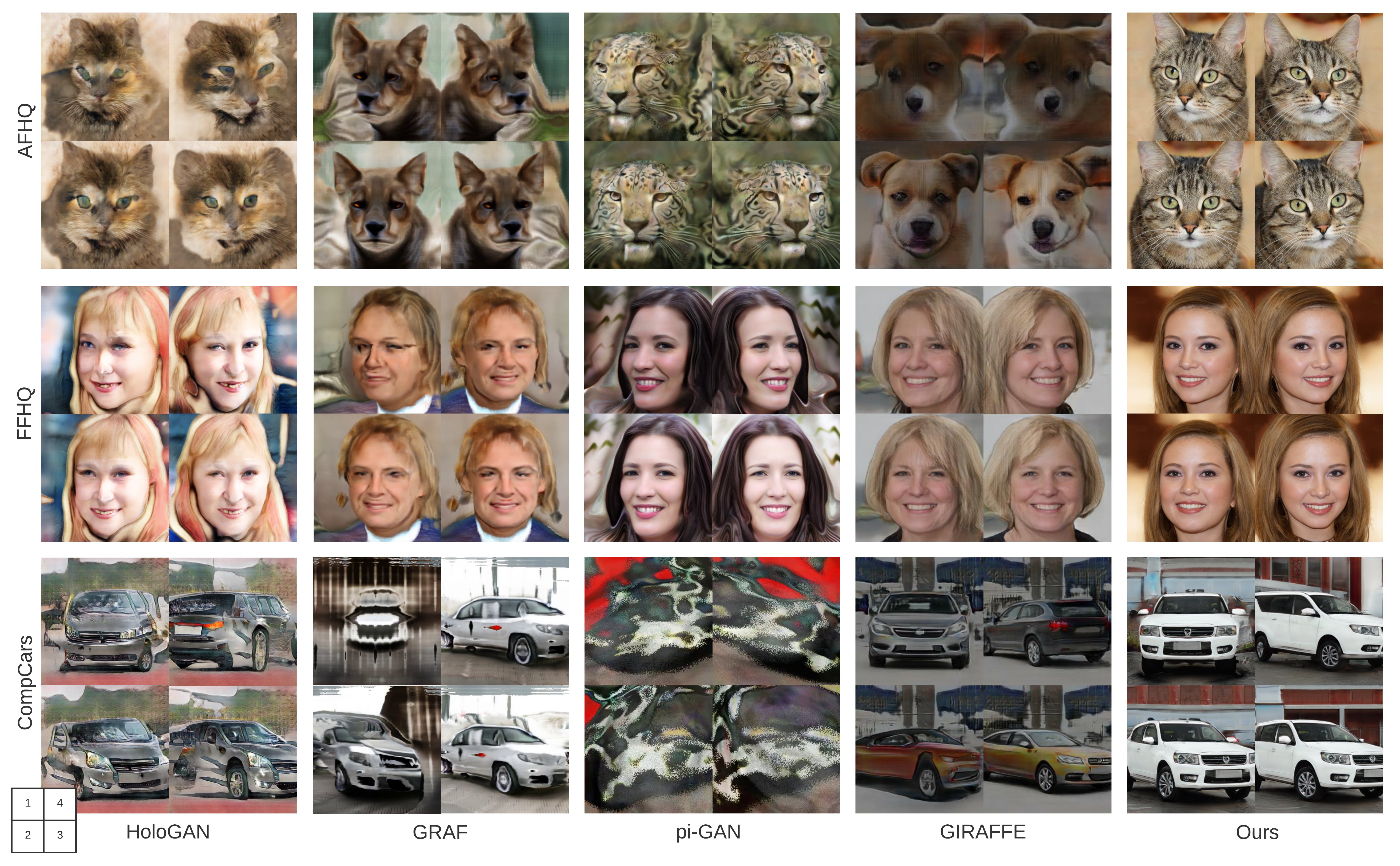}
    \caption{Qualitative comparisons at $256^2$. \emph{\emph{StyleNeRF}} achieves the best quality and 3D consistency.}
    \label{fig:example}
    \vspace{-10pt}
\end{figure}
\section{Experiments}
\subsection{Experimental Settings}
\vspace{-5pt}\paragraph{Datasets}
We evaluate \emph{StyleNeRF} on four high-resolution unstructured real datasets: FFHQ~\citep{karras2019style}, MetFaces~\citep{karras2020training}, AFHQ~\citep{choi2020stargan} and CompCars~\citep{yang2015large}. % as well as the synthetic CARLA~\citep{schwarz2020graf}.
The dataset details are described in \Cref{sec.dataset}.
% \Cref{sec.dataset} describes the dataset details including statistics and pre-processing.
\vspace{-5pt}\paragraph{Baselines}
We compare our approach with a voxel-based method,  HoloGAN~\citep{nguyen2019hologan}, and three radiance field-based methods: GRAF~\citep{schwarz2020graf}, $\pi$-GAN~\citep{chan2021pi} and GIRAFFE~\citep{niemeyer2021giraffe}. 
As most of the baselines are restricted to low resolutions, we made the comparison at $256^2$ pixels for fairness. 
We also report the results of the state-of-the-art 2D GAN~\citep{karras2020analyzing} for reference. 
See more details in \Cref{sec.baseline}.
% \vspace{-5pt}\paragraph{Metrics}
% To quantify the synthesis quality, we report the Frechet Inception Distance~\citep[FID,][]{heusel2017gans} and Kernal Inception Distance~\citep[KID,][]{binkowski2018demystifying}. % and Inception score~\citep[IS,][]{salimans2016improved} for comparison.
\vspace{-5pt}\paragraph{Configurations} 
All datasets except MetFaces are trained progressively. However, the MetFaces dataset is too small to train stably, so we finetune from the pretrained model for FFHQ at the highest resolution.
% For all datasets except MetFaces, we use the same architecture regardless of resolutions. MetFaces is too small to train stably. Instead, we finetune from the FFHQ checkpoint directly at the highest resolution.
By default, we train $64$ images per batch, and set $T_1=500k,T_2=5000k$ and $T_3=25000k$ images, respectively. 
The input resolution is fixed $32^2$ for all experiments.
% For all datasets except MetFaces, we adopt the same architecture regardless of resolutions. MetFaces is too small to train stably. Instead, we finetune from the FFHQ checkpoint directly at the highest resolution.
% By default, we train $64$ images per batch, and set $T_1=500,T_2=5000$ and $T_3=25000$ kimgs, respectively. 
% The input resolution is fixed $32^2$ for all experiments.
\begin{wrapfigure}{R}{0.56\textwidth}
%\begin{figure}[t]
    \centering
    \vspace{-13pt}
    \includegraphics[width=\linewidth]{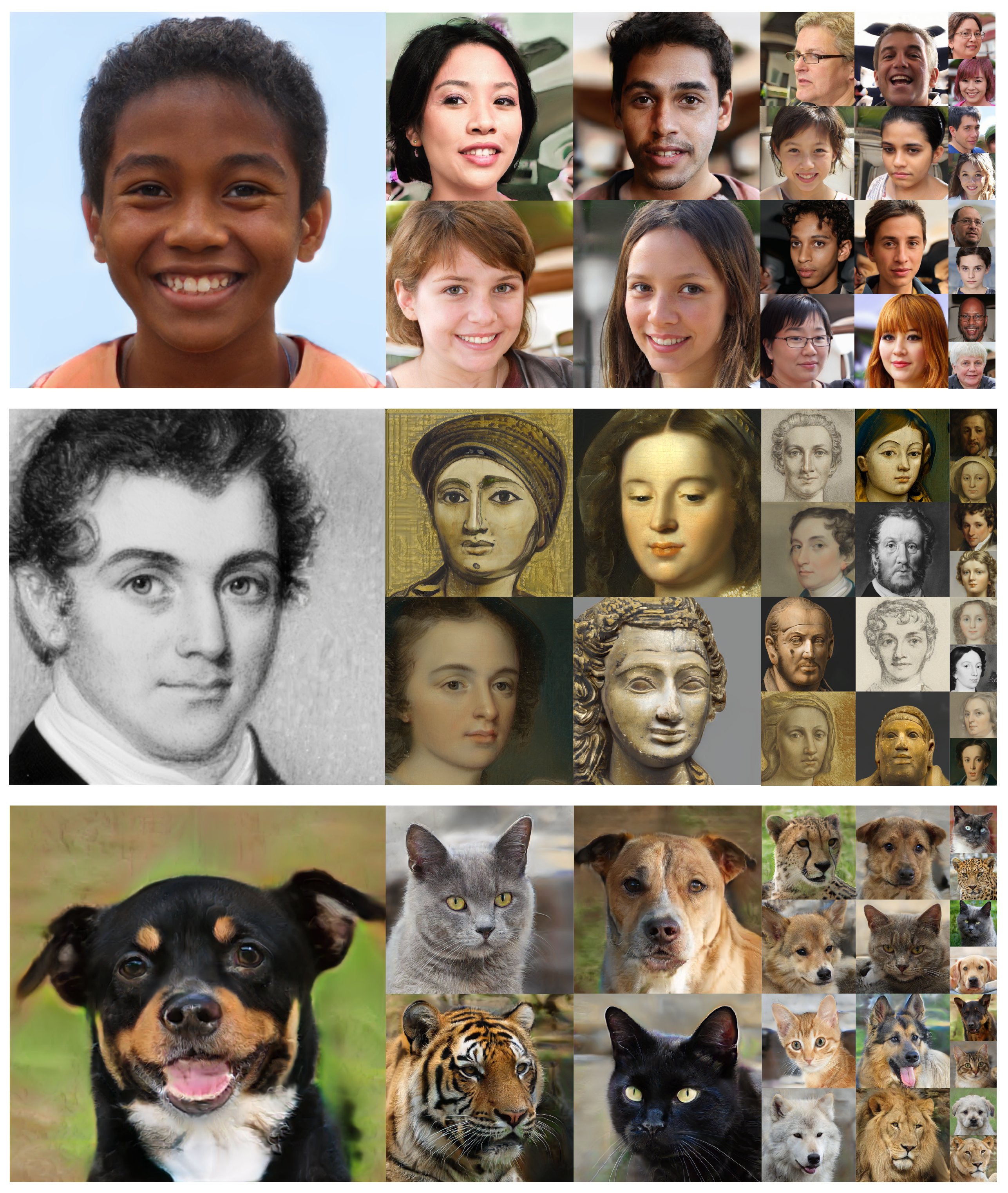}
    \caption{Uncurated set of images at $512^2$ produced by \emph{\emph{StyleNeRF}} from three datasets. Each example is rendered from a randomly sampled camera.}
    \label{fig:random_example}
    \vspace{-16pt}
%\end{figure}
\end{wrapfigure}

\subsection{Results}
\paragraph{Qualitative comparison} We evaluate our approach and the baselines on three datasets: FFHQ, AFHQ, and CompCars. Target images are resized to $256^2$. We render each object in a sequence and sample four viewpoints shown counterclockwise in \Cref{fig:example}. % We keep the viewpoints roughly comparable across approaches. 
While all baselines can generate images under direct camera control, HoloGAN, GARF, and $\pi$-GAN fail to learn geometry correctly and thus produce severe artifacts. % \LJ{no description on HoloGAN}
GIRAFFE synthesizes images in better quality; however, it produces 3D inconsistent artifacts: the shape and appearance in the output change constantly when the camera moves.
We believe it is due to the wrong choice of $3 \times 3$ \texttt{Conv} layers.
% We believe it is due to the improper use of view direction and CNN renderer \LJ{CNN renderer: the wrong choice of $3 \times 3$ \texttt{Conv} layers}, which capture the dataset bias.
%on CompCars changes the car orientation and color constantly when the scene rotates. 
Compared to the baselines, \emph{StyleNeRF} achieves the best visual quality with high 3D consistency across views.
Please see more results in the supplemental video.
\vspace{-5pt}\paragraph{Quantitative comparison} 
We measure the visual quality of image generation by the Frechet Inception Distance~\citep[FID,][]{heusel2017gans} and Kernal Inception Distance~\citep[KID,][]{binkowski2018demystifying}
in~\Cref{tab.quality_compare}. Across all three datasets, \emph{StyleNeRF} consistently outperforms the baselines by significant gains in terms of FID and KID and largely reduces the performance gap between the 3D-aware GANs and the SOTA 2D GAN (i.e., StyleGAN2). Note that while 2D GANs can achieve high quality for each image, they cannot synthesize images of the same scene with 3D consistency. % and often do not allow explicit camera control. % which indicates \emph{StyleNeRF} has better generalization and modeling capability.
\begin{table}[t]
\begin{center}
\small
\caption{\label{tab.quality_compare} Quantitative comparisons at $256^2$. We calculate FID, KID$\times 10^3$, and the average rendering time (batch size $=1$). The 2D GAN (StyleGAN2~\citep{karras2020analyzing}) results are for reference.  }
% \scalebox{0.88}{
\begin{tabular}{l | cc cc cc  | rrrrr}
\toprule
  & \multicolumn{2}{c}{FFHQ $256^2$}  &  \multicolumn{2}{c}{AFHQ $256^2$} & \multicolumn{2}{c|}{CompCars $256^2$} & \multicolumn{5}{c}{Rendering time (ms / image)} \\
  Models & FID & KID & FID & KID & FID & KID & 64 & 128 & 256 & 512 & 1024\\
\midrule
 2D GAN & 4 & 1.1 & 9& 2.3& 3 & 1.6& -  & - & 46 & 51 & 53\\
% 2D GAN & 4.0 & 1.1 & 9.4& 2.3& 3.4 & 1.6& -  & - & 46 & 51 & 53\\
 \midrule
 HoloGAN  & 75 & 68.0 & 78 & 59.4 & 48& 39.6 & 213 & 215 & 222 & - & -\\
 GRAF  & 71 & 57.2 & 121 & 83.8 & 101 & 86.7 & 61 & 246 & 990 & 3852 & 15475\\
 $\pi$-GAN  & 85 & 90.0 & 47& 29.3 & 295 & 328.9 & 58 & 198 & 766 & 3063 & 12310\\
%  $\pi$-GAN  & 85.2 & 90.0 & 46.7& 29.3 & & & 58 & 198 & 766 & 3063 & 12310\\
% GIRAFFE & 34.5 & 23.7 & & & & & 8 & - & 9 & - & -\\
GIRAFFE & 35 & 23.7 & 31 &13.9 & 32& 23.8& 8 & - & 9 & - & -\\
 \midrule
% Ours & 8.4 & 3.7 & 13.6 & 3.5& 8.5 & 4.3 & -& - & 65 & 74 & 98\\
Ours & 8 & 3.7 & 14 & 3.5& 8 & 4.3 & -& - & 65 & 74 & 98\\
\bottomrule
\end{tabular}
%}
\end{center}
\vspace{-10pt}
\end{table}
% \vspace{-5pt}\paragraph{Speed comparison}
\Cref{tab.quality_compare} also shows the speed comparison over different image resolutions on FFHQ. \emph{StyleNeRF} enables rendering at interactive rates and achieves significant speed-up over voxel and pure NeRF-based methods, and is comparable to GIRAFFE and StyleGAN2. % \JG{GIRAFFE is even faster but }
\vspace{-5pt}
\paragraph{High-resolution synthesis} 
Unlike the baseline models, our method can generate high-resolution images ($512^2$ and beyond). 
\Cref{fig:random_example} shows an uncurated set of images rendered by \emph{StyleNeRF}.
We also show more results of \emph{\emph{StyleNeRF}} and report the quantitative results of the SOTA 2D GAN (StyleGAN2~\citep{karras2020analyzing}) for reference in the \Cref{sec.additional_results}.
\begin{figure}[t]
    \centering
    \includegraphics[width=\linewidth]{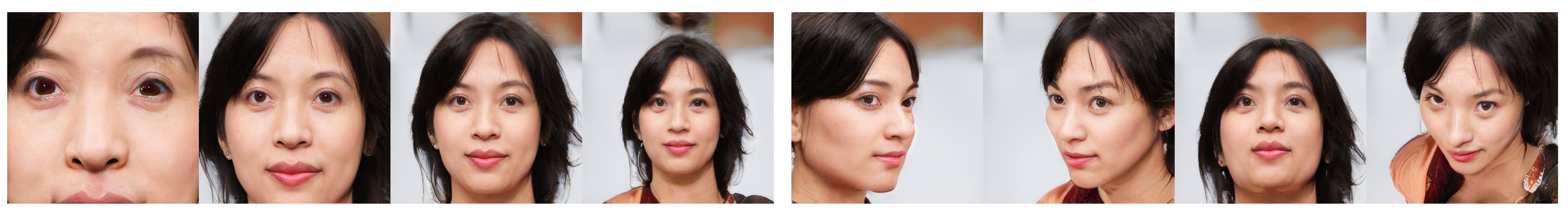}
    \caption{Images synthesized from camera poses which starkly differ from training camera poses. }
    \vspace{-10pt}
    \label{fig:extreme}
\end{figure}
\vspace{-5pt}

\subsection{Controllable Image Synthesis}
% In this section, we show how we can explicitly control a pretrained \emph{StyleNeRF} model.
% \paragraph{Camera control via extreme poses} 
\vspace{-5pt}\paragraph{Explicit camera control} 
%As the NeRF backbone encodes the 3D information, 
Our method can synthesize novel views with direct camera control and generalize to extreme camera poses, which starkly differs from the training camera pose distribution. \Cref{fig:extreme} shows our results with extreme camera poses, such as zoom-in and -out, and steep view angles.
The rendered images maintain good consistency given different camera poses.
%Though some artifacts appear on the neck and the shoulders when extreme view angles are used due to the rare corresponding training cameras, the central face still looks plausible.

\begin{figure}[t]
    \centering
    \includegraphics[width=\linewidth]{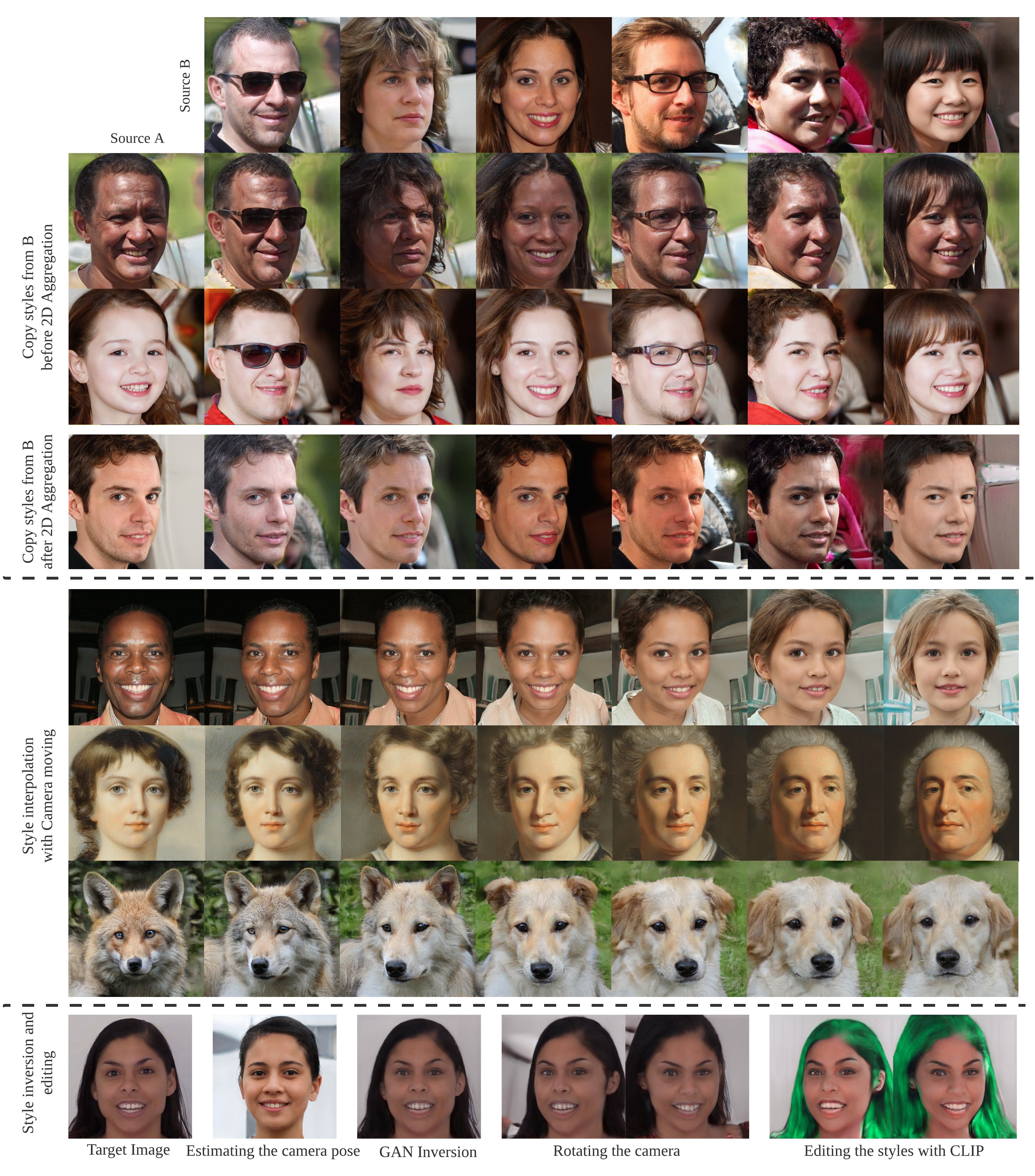}
    \caption{\emph{Style mixing (Top)}: % two sets of images were generated from their latent codes (sources A and B), and 
    images were generated by copying the specified styles from source B to source A. All images are rendered from the same camera pose. \emph{Style interpolation (Middle)}: we linearly interpolate two sets of style vectors (leftmost and rightmost images) while rotating the camera.
    \emph{Style inversion and editing (Below)}: the target image is selected from the DFDC dataset~\citep{dolhansky2019deepfake}. To edit with CLIP scores, we input ``a person with green hair'' as the target text.
    % For style inversion and editing, the target image is selected from the FFHQ dataset. To edit with CLIP, we input ``a person with purple hair'' and ``orange hair'', respectively.
    %For mixing experiments, we copy the styles before or after 2D aggregation from source B to the styles of source A.
    }
    \label{fig:style_mix_interp}
    \vspace{-10pt}
\end{figure}

\vspace{-5pt}\paragraph{Style mixing and interpolation}
% It is the same experiments in StyleGAN, mixing styles and interpolate styles. 
% In standard StyleGAN (\cite{karras2019style}), it has been shown by performing style-mixing and -interpolation experiments that the style vector can encode rich high-level attributes of appearances. ,
\Cref{fig:style_mix_interp} shows the results of style mixing and interpolation.
As shown in style mixing experiments, copying styles before 2D aggregation affects geometry aspects (shape of noses, glasses, etc.), while copying those after 2D aggregation brings changes in appearance (colors of skins, eyes, hairs, etc.), which indicates clear disentangled styles of geometry and appearance.
In the style interpolation results, the smooth interpolation between two different styles without visual artifacts further demonstrates that the style space is semantically learned. % demonstrates the high-level meaningfulness of the learned style space \LJ{demonstrates... -> demonstrates that the style space is learned well?}.

% \begin{figure}[t]
%     \centering
%     \vspace{-5pt}
%     \includegraphics[width=\linewidth]{figures/inversion}
%     \caption{Style inversion and editing. The target image is selected from the FFHQ dataset. To edit with CLIP, we input ``a person with purple hair'' and ``orange hair'', respectively.}
%     \vspace{-10pt}
%     \label{fig:inversion}
% \end{figure}
% \paragraph{Levels of Styles}

\begin{wrapfigure}{R}{0.44\textwidth}
%\begin{figure}[t]
    \centering
    \vspace{-13pt}
    \includegraphics[width=\linewidth]{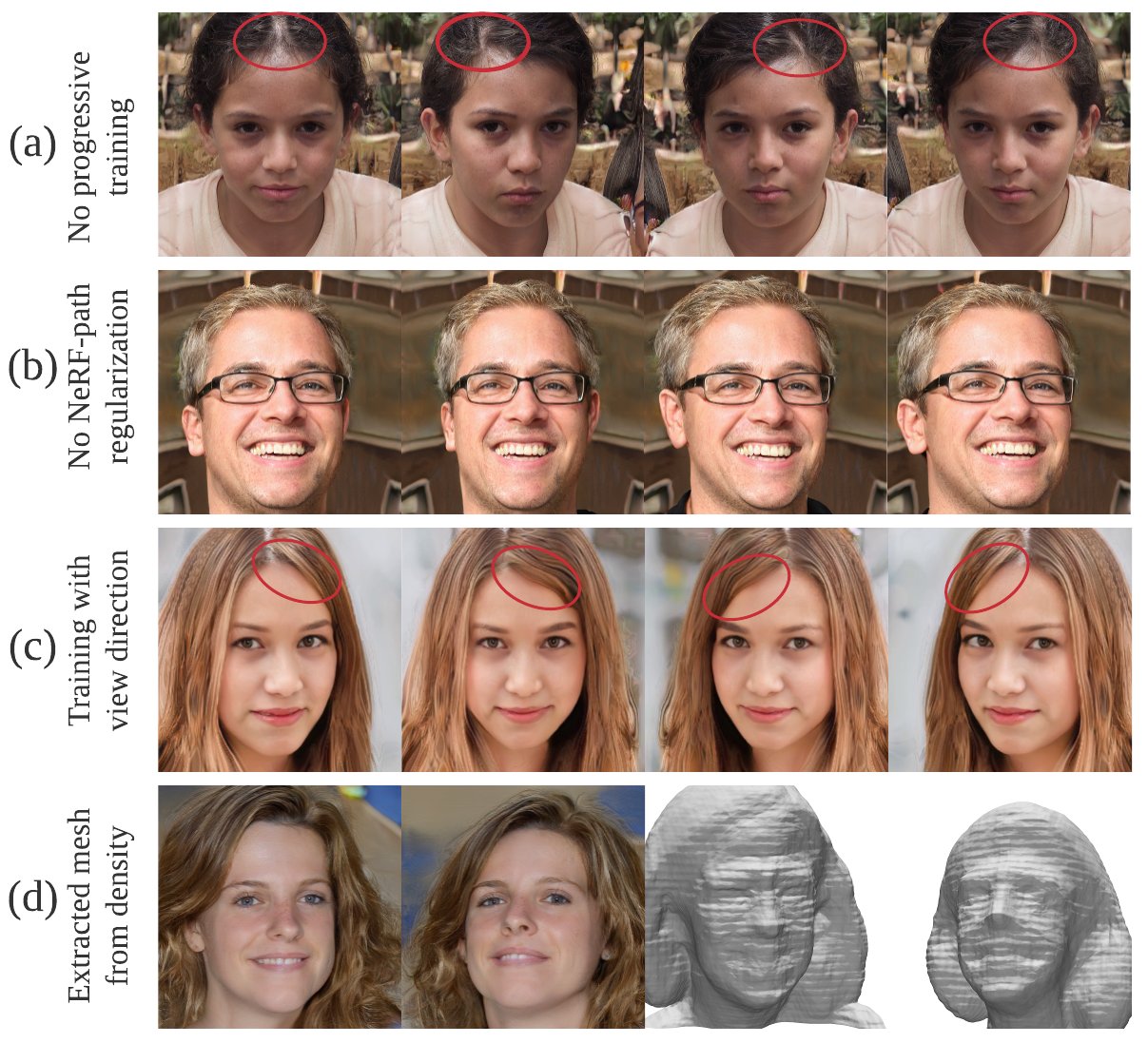}
    \caption{Failure results of ablation studies and limitations.}
    \label{fig:ablation}
    \vspace{-30pt}
%\end{figurei
\end{wrapfigure}
\vspace{-5pt}
\paragraph{Style inversion and editing} %To further demonstrate the power of the learned style space of \emph{StyleNeRF}, 
% Following \cite{karras2020analyzing}, 
We can also take a learned \emph{StyleNeRF} model for inverse rendering tasks. To achieve this, we first pre-train a camera pose predictor in a self-supervised manner.
Given a target image, we use this predictor to estimate the camera pose and optimize the styles via back-propagation
to find the best styles that match the target image. % The resulting styles can be rendered free from novel viewpoints. 
% we use the trained model to conduct inverse rendering tasks as in  - finding the latent style vector to re-synthesis an image given the real counterpart. To achieve it, we optimize the style vector by minimizing the loss function defined in Pixel2Style2Pixel (\cite{richardson2021encoding}) {\color{red} to check}, while the camera angle is predicted by a pre-trained image encoder.
After obtaining the styles, we can further perform semantic editing. For instance, we can optimize the styles according to text snippet using a CLIP loss~\citep{radford2021learning,patashnik2021styleclip}. An example is shown in \Cref{fig:style_mix_interp}.

\subsection{Ablation Studies}

% \paragraph{Internal representations with different upsampling}
% \paragraph{Background style}
%\paragraph{NeRF part size}
%\vspace{-5pt}\paragraph{Importance of progressive training}
\Cref{fig:ablation}~(a) shows a typical result without progressive training. Despite not directly affecting the quality of each single image, the model fails to learn the correct shape, e.g. the face should be in convex shape but is predicted as concave surfaces. This leads to severe 3D inconsistent artifacts when the camera moves (see the red circles in \Cref{fig:ablation}~(a)) % \LJ{add red circles}. 
% \vspace{-5pt}\paragraph{Importance of regularization}

As shown in \Cref{fig:ablation}~(b), without NeRF-path regularization, the model sometimes gets stuck learning geometry in 3D and produces a ``flat shape" output. % \LJ{why this happened?}
% We show the results without using NeRF-path regularization in \Cref{fig:ablation}~(b). Without the guidance of the NeRF path, the model gets stuck and produces a ``flat shape'' output. In such a case, the camera will fail to control the rotation of the output image.
% \vspace{-5pt}\paragraph{Importance of removing view direction}

\Cref{fig:ablation}~(c) demonstrates an example of the result with view direction as a condition.
The modeling of view-dependent effects introduces an ambiguity between 3D shape and radiance and thus leads to a degenerate solution without multi-view consistency (see the red circles in \Cref{fig:ablation}~(c)). 
% The model uses the view information to make up hair orientation when the viewpoint moves, resulting in inconsistent visualization.

\subsection{Limitations and Future work}
% worse geometry comparing to the pure NeRF based method
While \emph{StyleNeRF} can efficiently synthesize photo-realistic high-resolution images in high multi-view consistency with explicit camera control, 
it sacrifices some properties that pure NeRF-based methods have. An example is shown in \Cref{fig:ablation}~(d), where we extract the underlying geometry with marching cube from the learned density. 
Although \emph{StyleNeRF} is able to recover a coarse geometry, it captures less  details compared to the pure NeRF-based models such as $\pi$-GAN (\cite{chan2021pi}). 
% In addition, while \emph{StyleNeRF} preserves multi-view consistency to some extent, there is no guarantee for exact consistency. 
Additionally, \emph{StyleNeRF} only empirically preserves multi-view consistency, but it does not guarantee strict 3D consistency. Further exploration and theoretical analysis are needed.

\section{Conclusion}
We proposed a 3D-aware generative model, \emph{StyleNeRF}, for efficient high-resolution image generation with high 3D consistency, which allows control over explicit 3D camera poses and style attributes. 
%\emph{StyleNeRF} introduced neural radiance fields (NeRF) into a style-based generative model and addressed two challenges in this problem, that is, improving rendering efficiency for high-resolution image synthesis, and preserving 3D consistency in image generation. The first issue has been addressed by employing volume rendering to produce only a low-resolution feature map and upsampling the feature map into high resolution. To enforce 3D consistency, we proposed several designs, including a desirable upsampling strategy and a new regularization term.
%a new per-pixel \LJ{add: 3D?} noise injection method. 
Our experiments have demonstrated that \emph{StyleNeRF} can synthesize photo-realistic $1024^2$ images at interactive rates and outperforms previous 3D-aware generative methods. 

%while highly preserving 3D consistency, which is not possible to be achieved both by previous methods. We also demonstrated the use of  \emph{StyleNeRF} on  zoom-in and -out, style mixing and interpolation, style inversion, and text-to-image editing.

\section*{Ethics Statement}

Our work focuses on technical development, i.e., synthesizing high-quality images with user control. Our approach can be used for movie post-production, gaming, helping artists reduce workload, generating synthetic data to develop machine learning techniques, etc. Note that our approach is not biased towards any specific gender, race, region, or social class. It works equally well irrespective of the difference in subjects.

However, the ability of generative models, including our approach, to synthesize images at a quality that some might find difficult to differentiate from source images raises essential concerns about different forms of disinformation, such as generating fake images or videos. 
Therefore, we believe the image synthesized using our approach must present itself as synthetic. We also believe it is essential to develop appropriate privacy-preserving techniques and large-scale authenticity assessment, such as fingerprinting, forensics, and other verification techniques to identify synthesized images. Such safeguarding measures would reduce the potential for misuse. 

We also hope that the high-quality images produced by our approach could foster the development of the forgery mentioned above detection and verification systems. Finally, we believe that a robust public conversation is essential to creating a set of appropriate regulations and laws that would alleviate the risks of misusing these techniques while promoting their positive effects on technology development.

\section*{Reproducibility Statement}

We assure that all the results shown in the paper and supplemental materials can be reproduced. Furthermore, we will open-source our code together with pre-trained checkpoints.
To reproduce our results, we provide the preprocessing procedures of each dataset and implementation details in the main paper and Appendix.

\section*{Acknowledgements} Christian Theobalt was supported by ERC Consolidator Grant 4DReply (770784). Lingjie Liu was supported by Lise Meitner Postdoctoral Fellowship.

\bibliography{paper_conference}
\bibliographystyle{paper_conference}
\appendix
\newpage

%\section{Progressive Training Details}
%In this work, we adopt a \textit{three}-stage approach for progressive growing:
%Let the input and target resolutions as $R_{\text{in}}$ and $R_{\text{out}}$, respectively. 
%for the first stage ($T_1$), we train StyleNeRF strictly without approximation at low-resolution; then, %new layers are faded linearly into 
%both the generator and discriminator linearly increases until reaching the target resolution ($T_2$); lastly, we continue training the model at highest resolution to convergence ($T_3$). Refer to Appendix for more details
%For the first stage ($T_1$), we train StyleNeRF strictly without approximation at low-resolution.

\section{Additional Implementation Details}
\label{sec.details}
\subsection{Implicit Fields Modeling}
\vspace{-5pt}\paragraph{Modeling backgournd with NeRF++}
Most unstructured images used in this work have a complex and unbounded background, especially for the images taken from $360$ degree directions (e.g., images in the CompCars dataset). Therefore, it is inefficient and difficult to model the entire embedded scene in the image within a fixed bounding box. Instead, following NeRF++~\citep{zhang2020nerf++}, we partition the whole scene into foreground and background where the background is modeled with an additional network that takes as inputs the background latent codes and a 3D point $\vx=(x, y, z)$ transformed using inverted sphere parameterization:
\begin{equation}
    \vx' = \left(x/r, y/r, z/r, 1/r\right), \ \ \text{where} \ \  r = \sqrt{x^2 + y^2 + z^2}
\end{equation}
Then, we uniformly sample $G$ background points in an inverse depth range: $[1/R, 0)$ where $R=2.0$ in our experiments to represent where the background starts.

\vspace{-5pt}\paragraph{Hierarchical volume sampling} Similar to NeRF~\citep{mildenhall2020nerf}, we adopt an efficient hierarchical sampling strategy for the foreground network learning. We first uniformly sample $N$ points between the near and far planes and then perform importance sampling of $M$ points based on the estimated density distribution in the coarse sampling stage. Different from NeRF~\citep{mildenhall2020nerf}, we use a single network for both the coarse and fine sampling to predict the color and density.

\subsection{Camera}
\vspace{-5pt}\paragraph{Camera setup}
We assume that each image is rendered by a camera with a fixed intrinsic matrix in a pose sampled from a pre-defined camera pose distribution.
The intrinsics are determined by field-of-view (FOV) and image resolution, normalized to $[-1,1]$ regardless of the actual resolution. The camera is located on the unit sphere, pointing at the world origin where the target object is located.
Then the camera pose $\bm{p}(\theta, \phi)$ is the function of pitch ($\theta$) and yaw ($\phi$). Typically, the cameras in real data do not follow the simple distributions that we assumed, which is the main problem for all 3D-aware GANs (including the proposed model) to learn the correct geometry. To tackle this issue, \citet{niemeyer2021campari} proposed to learn a neural network to generate camera simultaneously poses jointly with scene representations. However, our initial attempts of applying similar methods to StyleNeRF did not work, and the camera generator quickly diverged. Instead, we manually set the parameters $\phi$ and $\theta$ from either a Gaussian or uniform distribution depending on the datasets. 
\begin{wrapfigure}{R}{0.56\textwidth}
%\begin{figure}[t]
    \centering
    \vspace{-13pt}
    \includegraphics[width=\linewidth]{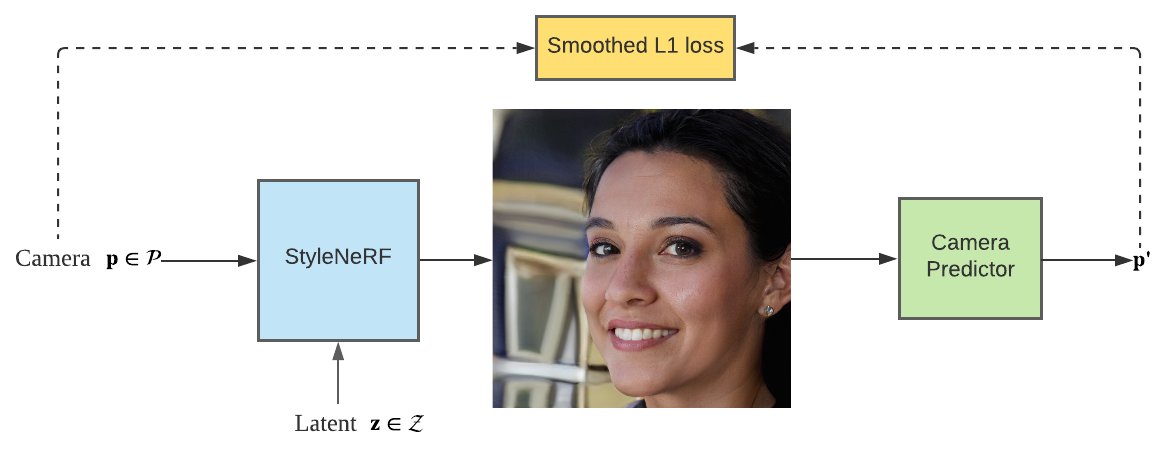}
    \caption{An illustration of self-supervised training the camera predictor.}
    \label{fig:camera}
    \vspace{-5pt}
%\end{figure}
\end{wrapfigure}
\vspace{-5pt}\paragraph{Camera predictor}
Once StyleNeRF is trained, we can optionally train an additional camera predictor in a self-supervised manner. More specifically, we first randomly sample a camera pose and the latent codes to render the output image. Then, a CNN-based encoder (backbone initialized with a pretrained ResNet-18) reads the image and predicts the input camera parameters based on a smoothed L1 loss function.
The camera predictor can be directly used on natural images, in particular for style inversion tasks (see \Cref{fig:style_mix_interp}) as our early exploration shows that it is non-trivial to train the camera pose from random through latent optimization. 

\subsection{Details about the noise injection}
\label{sec.noise}
As mentioned in the paper, injecting per-pixel noise improves the generation quality of GANs. However, naively adding such noise as in~\citet{karras2019style} is not feasible for our purpose because such noise exists in the pixel space, and it does not move when the viewpoint changes. On the other hand, the output gets noisy if we inject random noise for each frame separately.
To avoid such inconsistency, the default setup in our experiments is to remove the noise injection as similarly done in~\citep{karras2021alias} which works reasonably but has less ability to model local stochastic variations, resulting in lower visual quality (in terms of metrics like FID, KID, etc).
\begin{figure}[t]
    \centering
    \includegraphics[width=\linewidth]{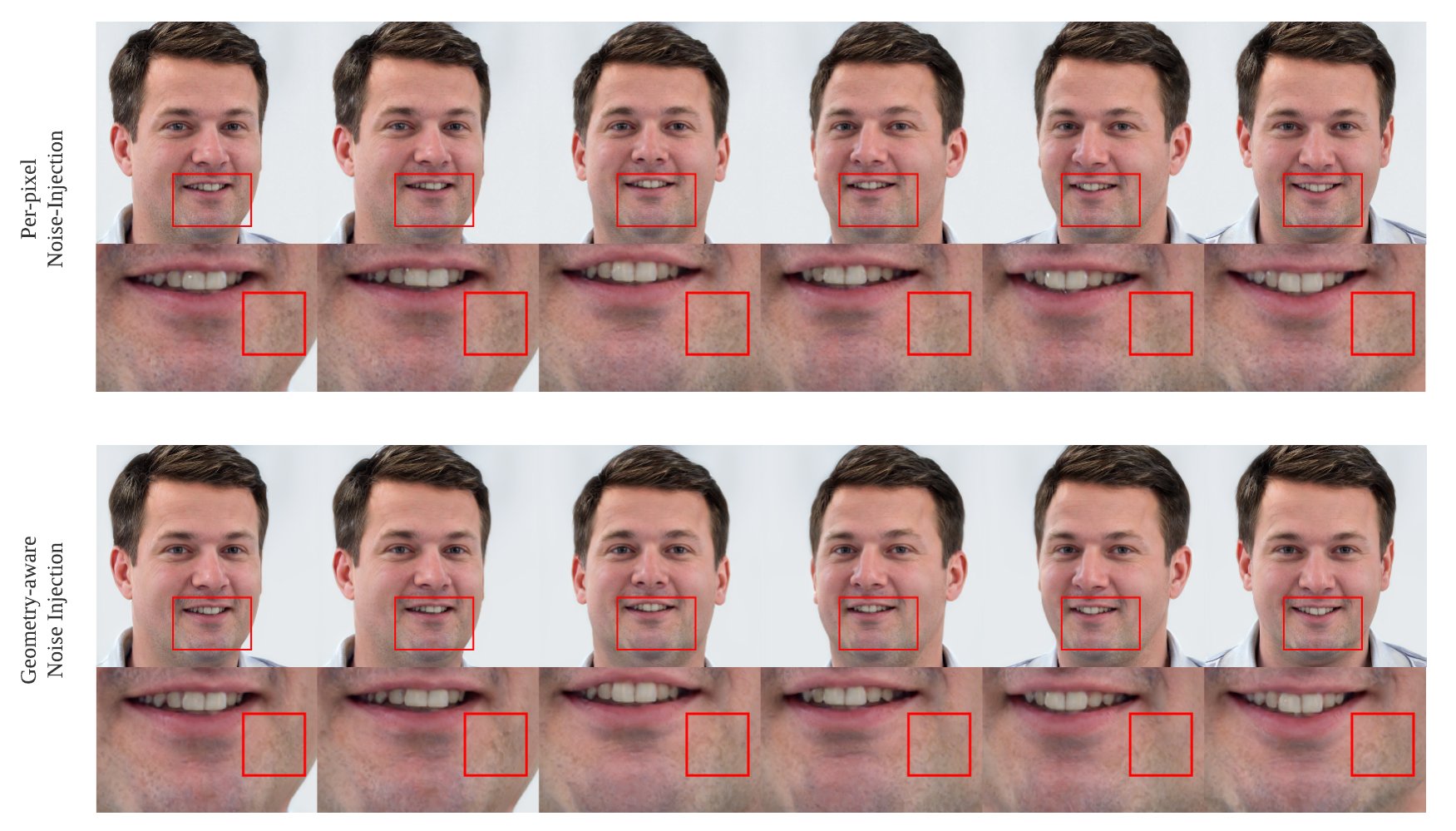}
    \caption{Example of sticking texture because of the injection of 2D noise at the test time. Note that the texture cropped inside the red box does not change across different views.}
    \vspace{-10pt}
    \label{fig:noise}
\end{figure}

As an alternative, to improve synthesizing quality and keep multi-view consistency, we design a geometry-aware method for noise injection. 
Precisely, for each feature layer at the resolution of $N^2$, we extract the underlying geometry using marching -cube based on the predicted density. Next, we set the volume resolution $N^3$ accordingly. Then, we assign independent Gaussian noise at each vertex of the extracted mesh and render the noise map via rasterization from the same viewpoint. Finally, we inject the resulting noise to each layer before non-linear activation. Since this process requires a proper density function to start with, we keep the standard per-pixel noise unchanged during training and only apply such geometry-aware noise at inference time. Compared to models without noise injection, although the proposed approach can capture more stochastic details, it is relatively slower (due to multi-scale marching-cube and rasterization), and the quality of the learned geometry constrains the final output.

\subsection{Progressive Training}
\label{sec.progressive}

As described in the main paper, it is essential to train StyleNeRF progressively from low to high resolutions to learn good geometry. We consider two types of progressive training:
\vspace{-5pt}\paragraph{Progressive Growing} A common approach is to follow \citet{karras2017progressive} where we start from a shallow network for low-resolution, and progressively fade in layers for high-resolutions. Since the parameters of the newly added layers are random, it is essential to have a linear interpolation with low-resolution output to stabilize training. %In such a case, however, we need to retrain a separate RGB layer for every resolution output.

\vspace{-5pt}\paragraph{Progressive Up-sampling}
Instead of growing new layers, another option is to train StyleNeRF by progressively inserting \texttt{upsample} operations during training as similarly done in \citet{chan2021pi}. In this case, we do not need to fade in layers linearly.

\begin{figure}[t]
    \centering
    \includegraphics[width=\linewidth]{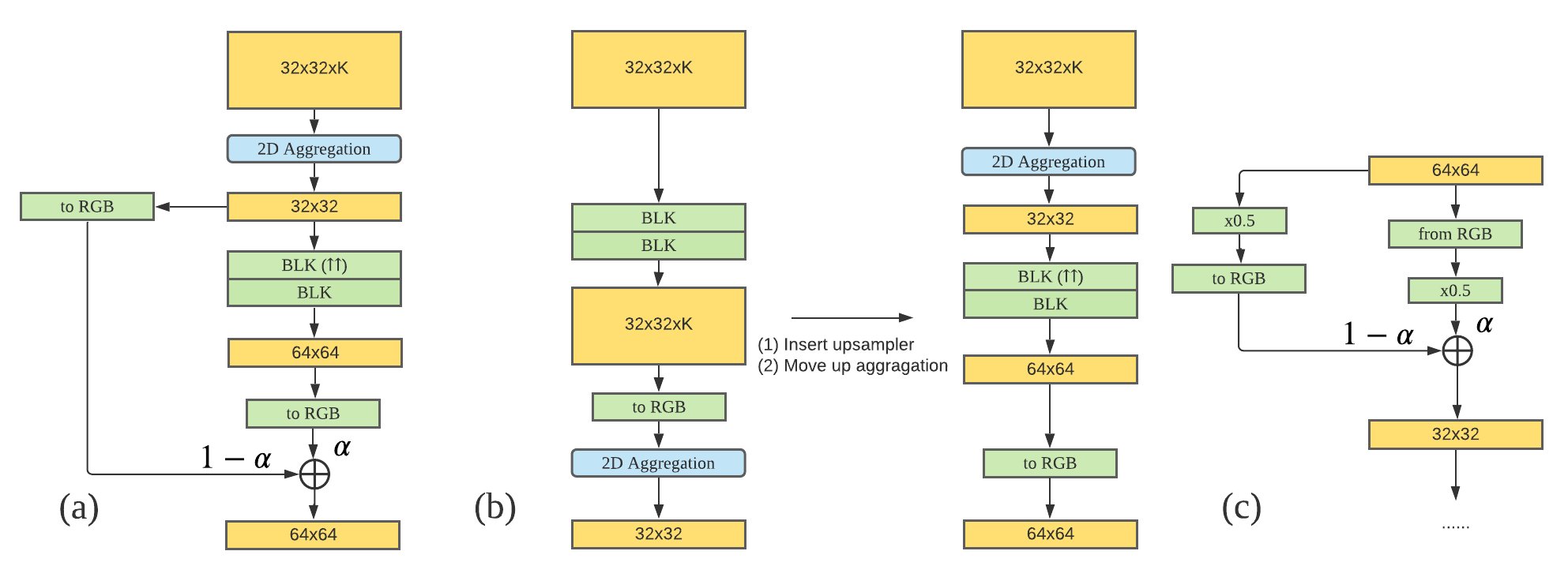}
    \caption{Illustrations of the proposed two variants of progressive training for StyleNeRF. (a) is for progressive growing of the generator; (b) is the same generator by progressively inserting upsamplers; (c) is the corresponding discriminator.}
    \label{fig:progressive}
\end{figure}

In both cases, the discriminator correspondingly grows from low to high resolutions progressively as described in ~\citet{karras2017progressive}. \Cref{fig:progressive} illustrates progressive training for the generator and the discriminator. In practice, we found that both methods worked similarly in terms of visual quality, while progressive growing achieves a better speed advantage for low-resolution images. Therefore, our results are reported by training with progressive growing.
\subsection{Hyperparameters}
We reuse the same architecture and default parameters of StyleGAN2 for the mapping network ($8$ fully connected layers, $100\times$ lower learning rate) and discriminator. In addition, both the latent and style dimensions are set to $512$.
For the foreground and background fields, before predicting the density, we use the style-based MLPs with $256$ and $128$ hidden units, respectively. 
The Fourier feature dimension is set to $L=10$ (\Cref{eq.pos}) for both fields.
For layers after 2D aggregation, we follow the flexible layer configuration used by StyleGAN2, which decreases the feature sizes from $512$ (for $32^2$) until $32$ (for $1024^2$).
We keep most of the training configurations unchanged from StyleGAN2.
\subsection{Implementation}
We implement our model based on the official Pytorch implementation of StyleGAN2-ADA~\footnote{\url{}https://github.com/NVlabs/stylegan2-ada-pytorch}. By default, we train the StyleNeRF model by going through $25000k$ images with a minibatch $=64$. All models are trained on 8 Tesla V100 GPUs for about three days.

\section{Dataset Details}
\label{sec.dataset}
\vspace{-5pt}\paragraph{FFHQ} (\url{https://github.com/NVlabs/ffhq-dataset}) contains 70k images of real human faces in resolution of $1024^2$. We assume the human face to be captured at the origin. In the training stage, we sample the pitch and yaw of the camera from the Gaussian distribution.

\vspace{-5pt}\paragraph{AFHQ} (\url{https://github.com/clovaai/stargan-v2#animal-faces-hq-dataset-afhq}) contains in total 15k images of animal faces including cat, dog and wild three categories in resolution of $512^2$. We directly merge all training images without using the label information. Similar to the FFHQ setting, we sample the pitch and yaw from the Gaussian distribution.

\vspace{-5pt}\paragraph{MetFaces} (\url{https://github.com/NVlabs/metfaces-dataset}) contains 1336 images of faces extracted from artworks. The original resolution is $1024^2$, and we resize it to $512^2$ for our main experiments. As mentioned in the paper, we finetune StyleNeRF from the FFHQ checkpoint due to the small size of MetFaces.

\vspace{-5pt}\paragraph{CompCars} (\url{http://mmlab.ie.cuhk.edu.hk/datasets/comp_cars/})  contains 136726 images capturing the entire cars with different styles. The original dataset contains images with different aspect ratios. We preprocess the dataset by center cropping and resizing them into $256^2$. In the training stage, we sample the camera $360$ degree uniformly.

\section{Baseline Details}
\label{sec.baseline}
\vspace{-5pt}\paragraph{HoloGAN~\citep{nguyen2019hologan}} We train HoloGAN on top of the official implementation \footnote{https://github.com/thunguyenphuoc/HoloGAN}. Specifically, HoloGAN is trained in resolution $256^2$ with Adam optimizer with an initial learning rate of 5e-5. We train each HoloGAN model for 50 epochs. The learning rate linearly decays between epoch 25 and 50. As original HoloGAN only supports training on images of resolution $64^2$ and $128^2$, to synthesize images of resolution $256^2$ we follow the same adaptation scheme as used in GIRAFFE~\citep{niemeyer2021giraffe} that add one convolution layer with AdaIn~\citep{huang2017arbitrary} and leaky ReLU activation on top of the official synthesis network.
\vspace{-5pt}\paragraph{GRAF~\citep{schwarz2020graf}} We adopt the official implementation~\footnote{https://github.com/autonomousvision/graf} and retrain GRAF models on FFHQ, AFHQ and preprocessed CompCars dataset in resolution $256^2$. Following the default setting, all models are trained at the target resolution using the patch-based discriminator.
\vspace{-5pt}\paragraph{$\pi$-GAN~\citep{chan2021pi}} Same as GRAF, we use the official implementation~\footnote{https://github.com/marcoamonteiro/pi-GAN} and retrain $\pi$-GAN models on these three datasets: FFHQ, AFHQ and CompCars. Due to the high cost of running $\pi$-GAN in high-resolution, we follow the same pipeline, which progressively increases the resolution from $32^2$ until $128^2$, and we render the final outputs in $256^2$ by sampling more pixels.
\vspace{-5pt}\paragraph{GIRAFFE~\citep{niemeyer2021giraffe}} For FFHQ dataset, we directly use their pretrained models of resolution $256^2$. For AFHQ and CompCars dataset, we retrain GIRAFFE with the official Giraffe implementation~\footnote{https://github.com/autonomousvision/giraffe}. We change the default random crop preprocessing to center crop to compare with other approaches.

\section{Additional Results}
\label{sec.additional_results}
\begin{figure}[t]
    \centering
    \includegraphics[width=\linewidth]{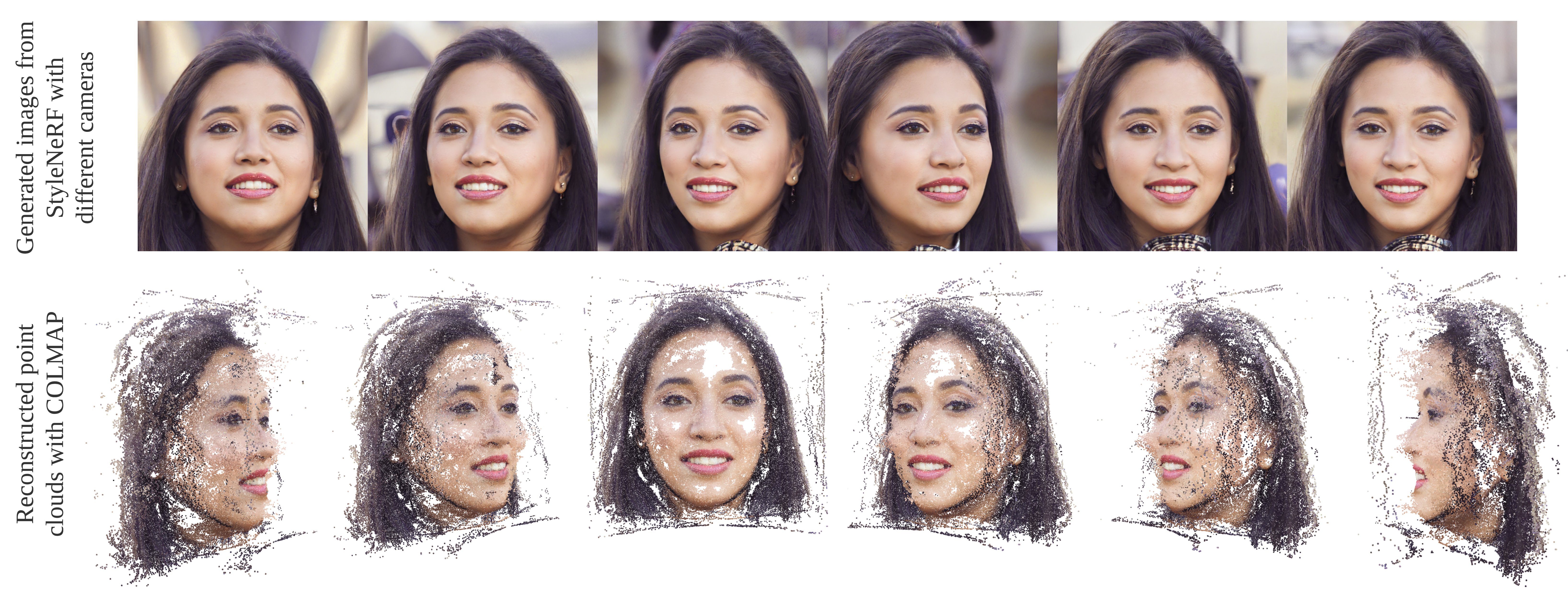}
    \caption{ COLMAP reconstructions for models trained on FFHQ $512^2$}
    \label{fig:colmap}
\end{figure}

We include additional visual results to show the quality and 3D consistency of the generated images. 
\vspace{-5pt}\paragraph{Comparison to StyleGAN2}
We also provide quantitative comparisons on high-resolution image synthesis against StyleGAN2 and a variant that adopts comparable settings with our default model. As shown in \Cref{tab.compare_stylegan},
there is still a gap to the best 2D-GAN models. Nevertheless, the proposed StyleNeRF reaches similar or even better results as 2D models in similar architectures. This implies the quality drops are mainly from less powerful architecture (e.g., no noise injection, no $3\times 3$ convolutions). As future work, more exploration can be done to close this gap.
\begin{table}[t]
\begin{center}
\small
\caption{\label{tab.compare_stylegan} (\emph{Left}) FID, KID $\times 10^3$ on images with $512^2$ and $1024^2$ pixels. 2D GAN represents the StyleGAN2~\citep{karras2020analyzing}. (\emph{Right}) We also compare the results of replacing $3\times 3$ \texttt{Conv} with $1\times 1$ and/or removing per-pixel noise injection for 2D models. We use \textbf{bold} font to represent the default setting for both our model.}
\begin{subtable}[t]{0.65\textwidth}
 \scalebox{0.88}{
\begin{tabular}[t]{lcccccccc}
\toprule
  \multirow{2}{*}{Models}& \multicolumn{2}{c}{FFHQ $512^2$}  &  \multicolumn{2}{c}{AFHQ $512^2$} & \multicolumn{2}{c}{MetFace $512^2$} & \multicolumn{2}{c}{FFHQ $1024^2$} \\
   & FID & KID& FID & KID & FID & KID& FID & KID\\
\midrule
 2D GAN & 3.1 & 0.7 & 8.6& 1.7 & 18.9& 2.7& 2.7 & 0.5 \\
%  \ \ w. $1\times1$ Conv, w/o noise injection & 16.4 & 3.6\\
% \midrule
 Ours & 7.8 & 2.2 & 13.2 & 3.6 & 20.4 & 3.3 & 8.1 & 2.4\\
%  \ \ w. noise injection & 6.0 & 1.6 \\
\bottomrule
\end{tabular}
}
\end{subtable}
\begin{subtable}[t]{0.34\textwidth}
 \scalebox{0.88}{
\begin{tabular}[t]{llcc}
\toprule
$1\times1$ & Noise& 2D GAN & Ours\\
\midrule
\texttt{Yes} & \texttt{No} & 11.4 & \bf{7.8} \\
\texttt{Yes} & \texttt{Yes}& 5.9 & 6.0\\
\texttt{No}  & \texttt{No} & 4.6 & - \\
\bottomrule
\end{tabular}
}
\end{subtable}
\end{center}
\vspace{-7pt}
\end{table}
\vspace{-5pt}\paragraph{3D Reconstruction}
We include a COLMAP~\citep{schoenberger2016sfm} reconstruction example of the proposed StyleNeRF on FFHQ to validate the 3D consistency of the model's output. As shown in \Cref{fig:colmap}, given a fixed style, we sample $36$ camera poses to generate images and obtain the reconstructed point clouds from COLMAP with default parameters and no known camera poses.

\vspace{-5pt}\paragraph{Additional Visual Results}
We show additional results of generating images with different camera views as follows. Please also refer to the supplemental video, which shows more results.

\begin{figure}
    \centering
    \includegraphics[width=\linewidth]{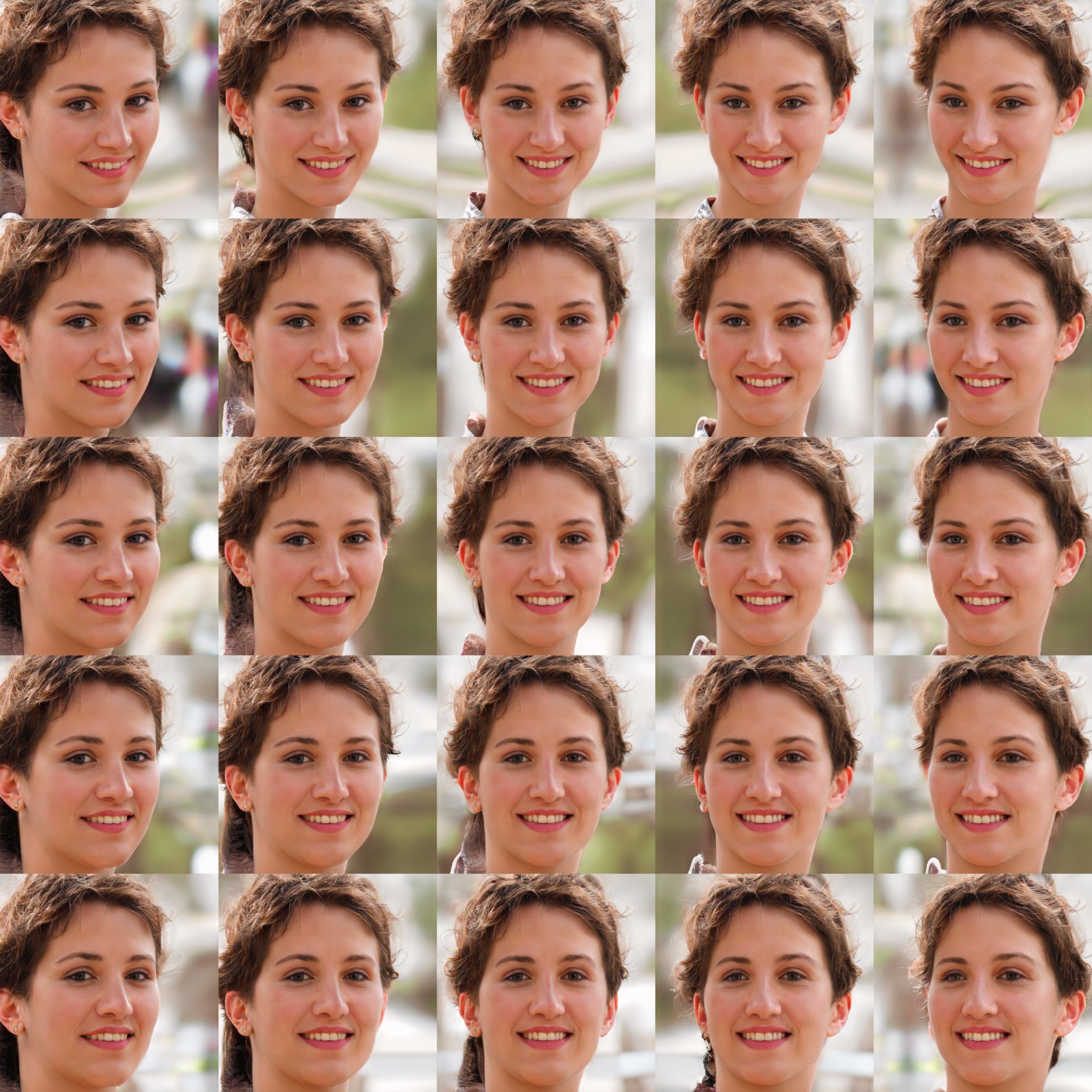}
    \caption{Example on FFHQ $1024^2$ displayed from different viewing angles.}
    \label{fig:grid_rotation}
\end{figure}
\begin{figure}
    \centering
    \includegraphics[width=\linewidth]{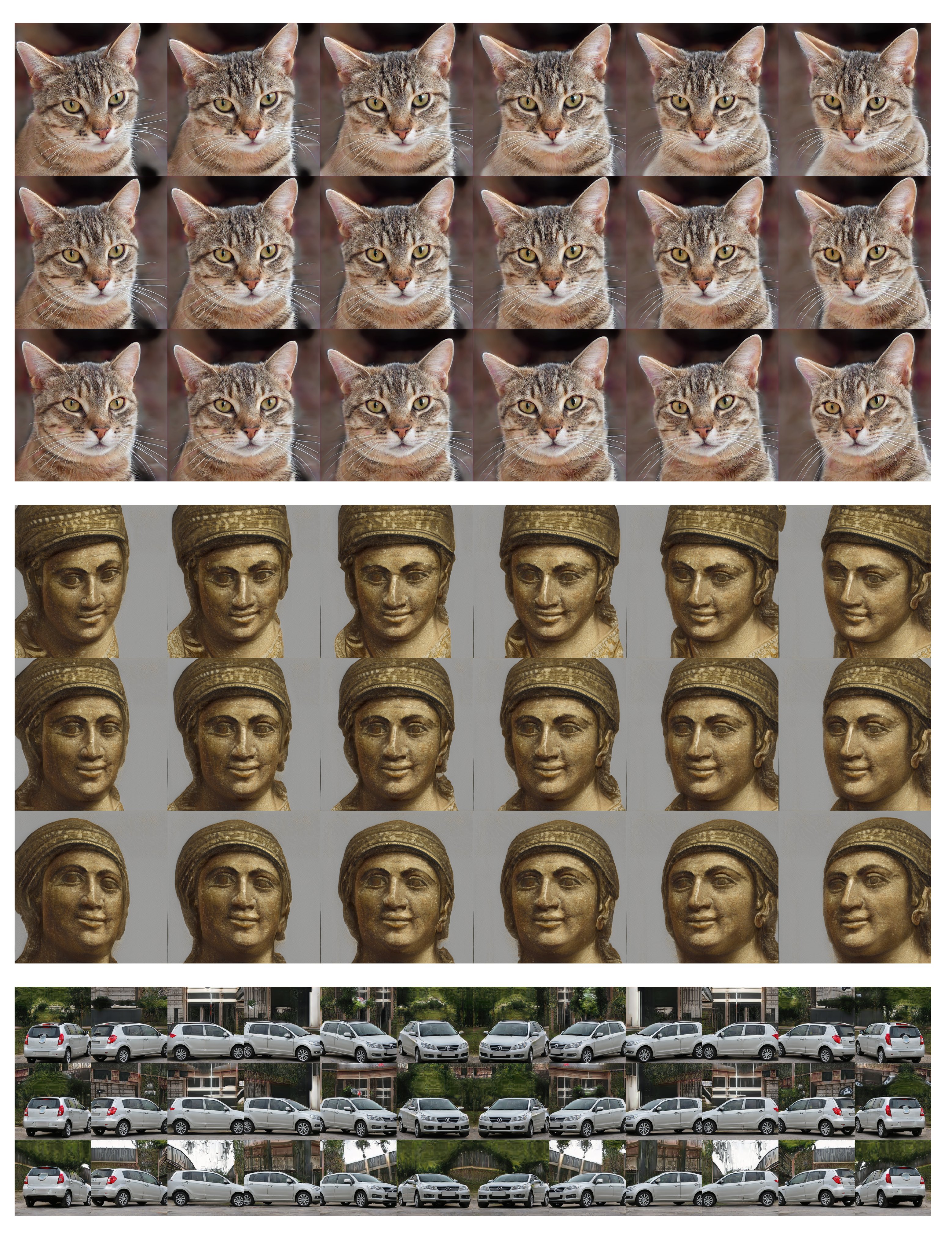}
    \caption{Example on AFHQ ($512^2$), MetFaces ($512^2$) and CompCars ($256^2$) displayed from different viewing angles.}
    \label{fig:grid_rotation}
\end{figure}
\begin{figure}
    \centering
    \includegraphics[width=\linewidth]{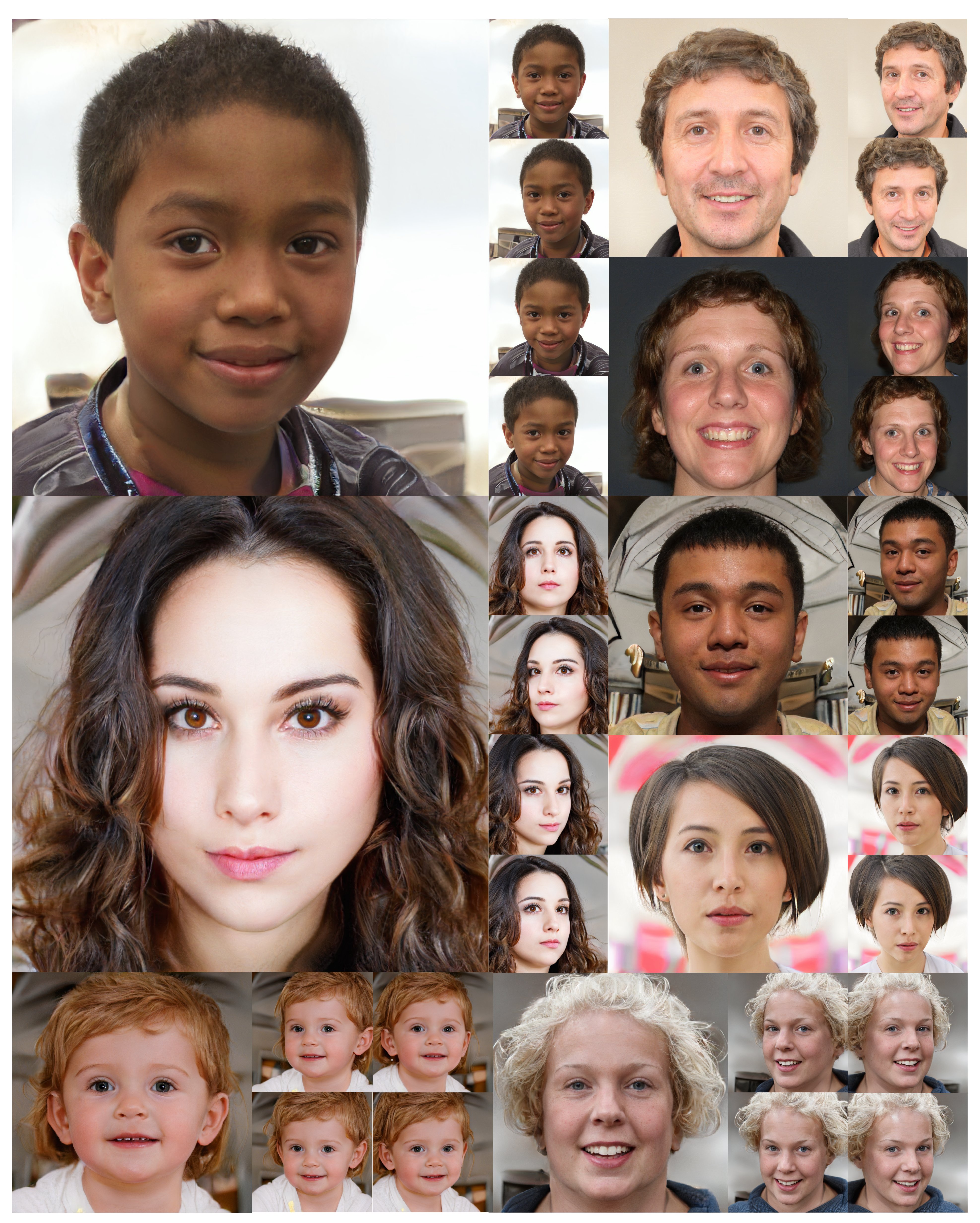}
    \caption{High resolution samples with explicit camera control on FFHQ $1024^2$.}
    \label{fig:high_res}
\end{figure}
\begin{figure}
    \centering
    \includegraphics[width=\linewidth]{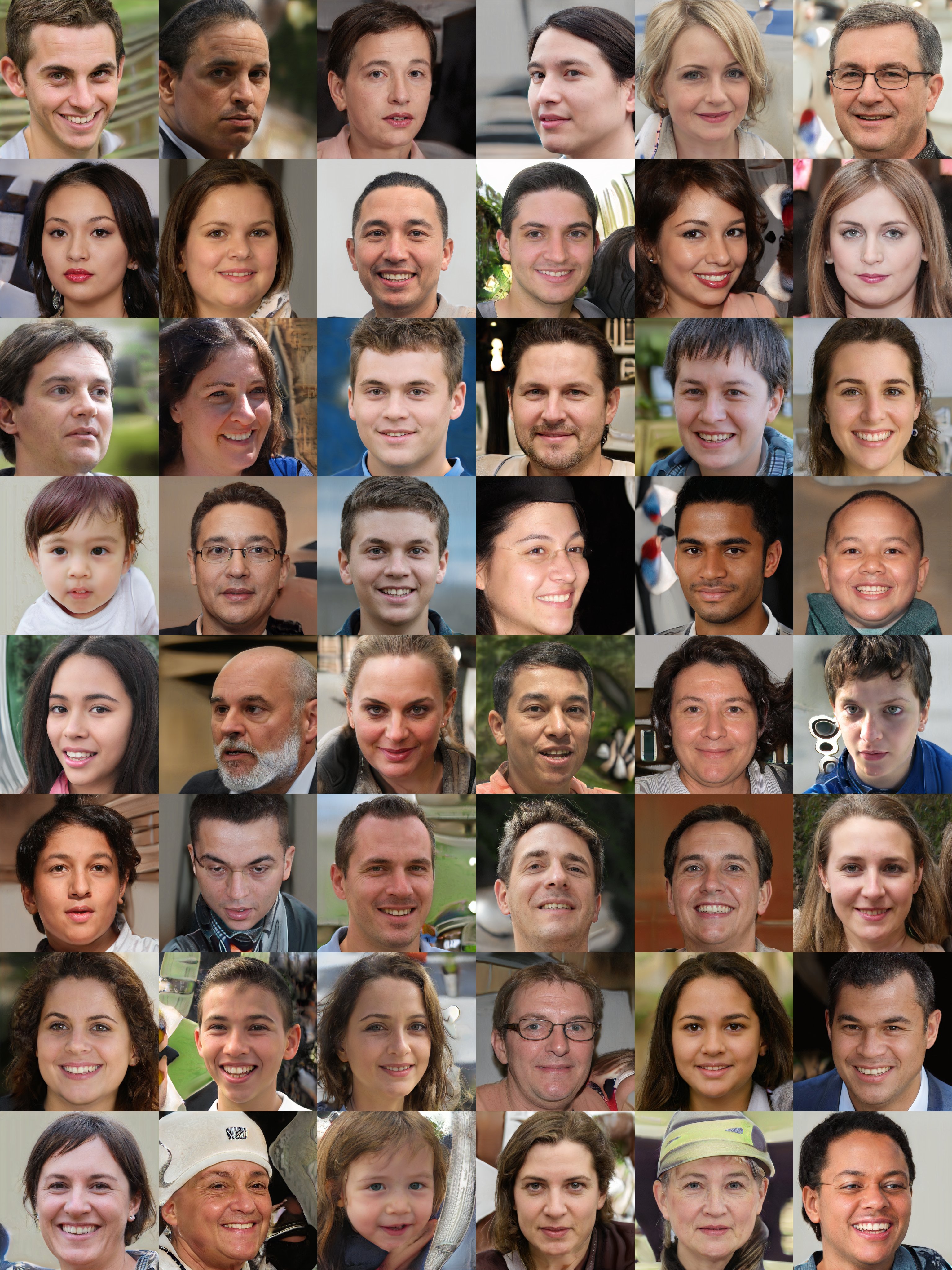}
    \caption{Random samples with random camera poses on FFHQ $512^2$.}
    \label{fig:random_metfaces}
\end{figure}
\begin{figure}
    \centering
    \includegraphics[width=\linewidth]{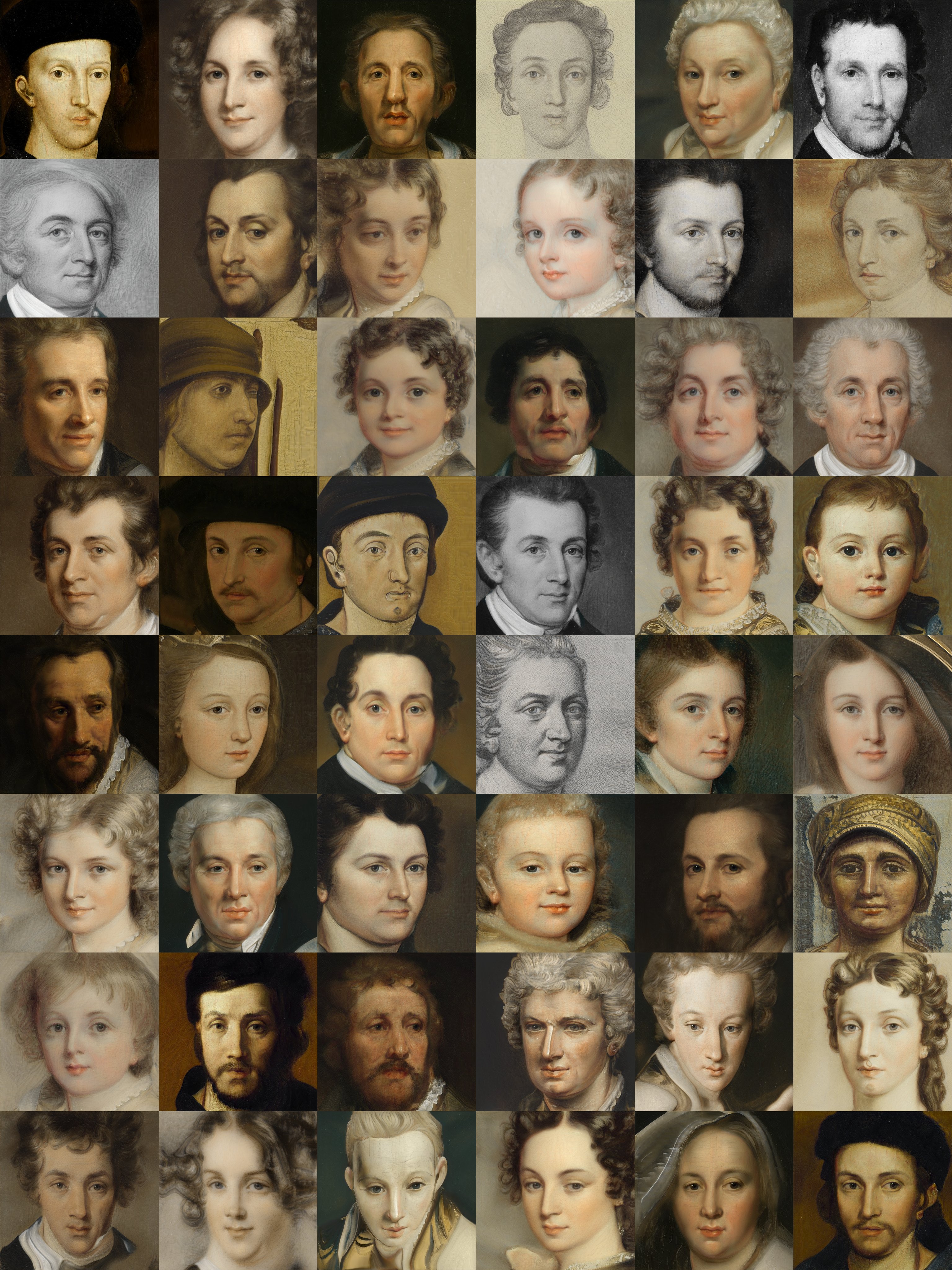}
    \caption{Random samples with random camera poses on MetFaces $512^2$.}
    \label{fig:random_metfaces}
\end{figure}
\begin{figure}
    \centering
    \includegraphics[width=\linewidth]{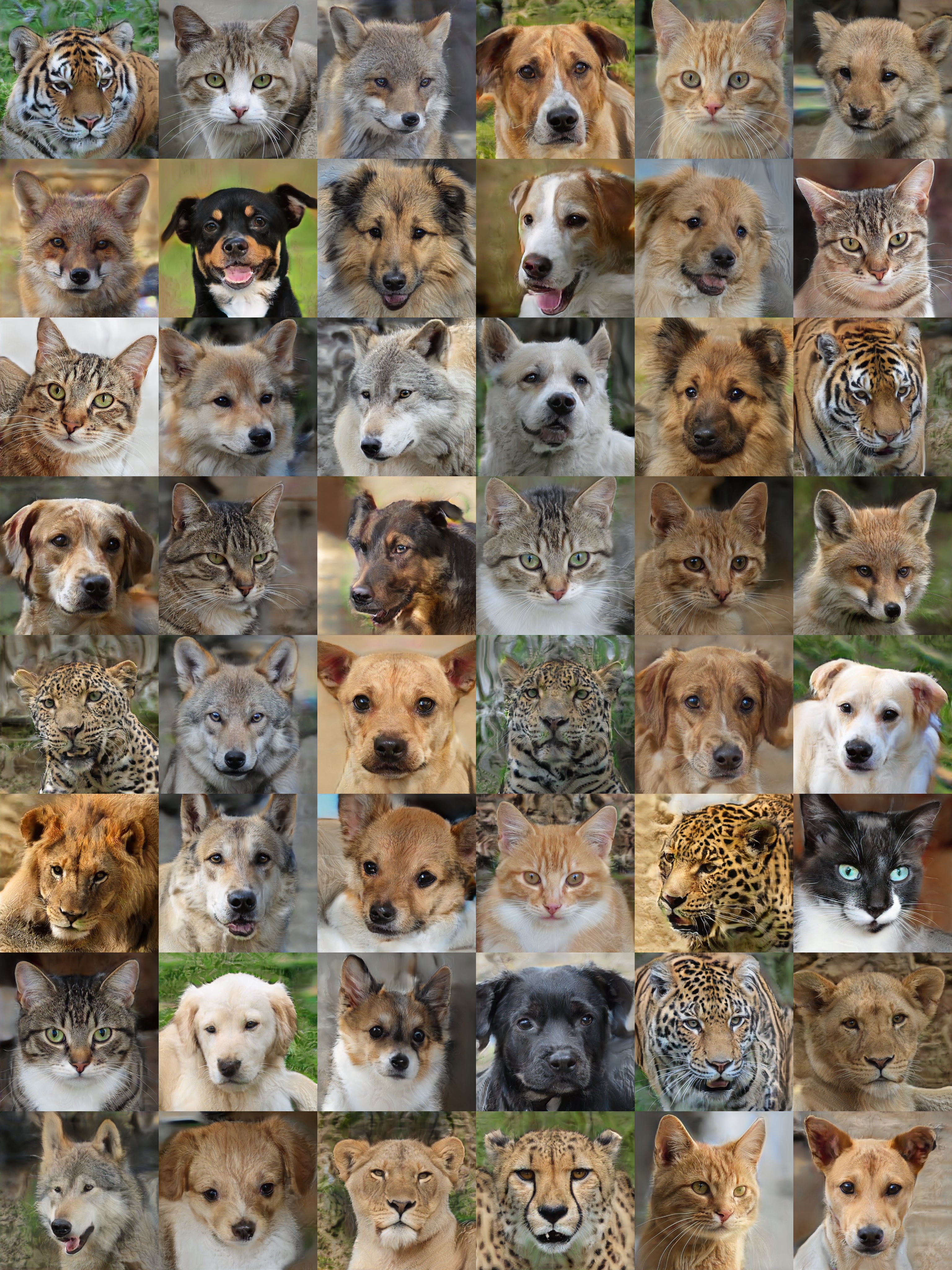}
    \caption{Random samples with random camera poses on AFHQ $512^2$.}
    \label{fig:random_afhq}
\end{figure}
\begin{figure}
    \centering
    \includegraphics[width=\linewidth]{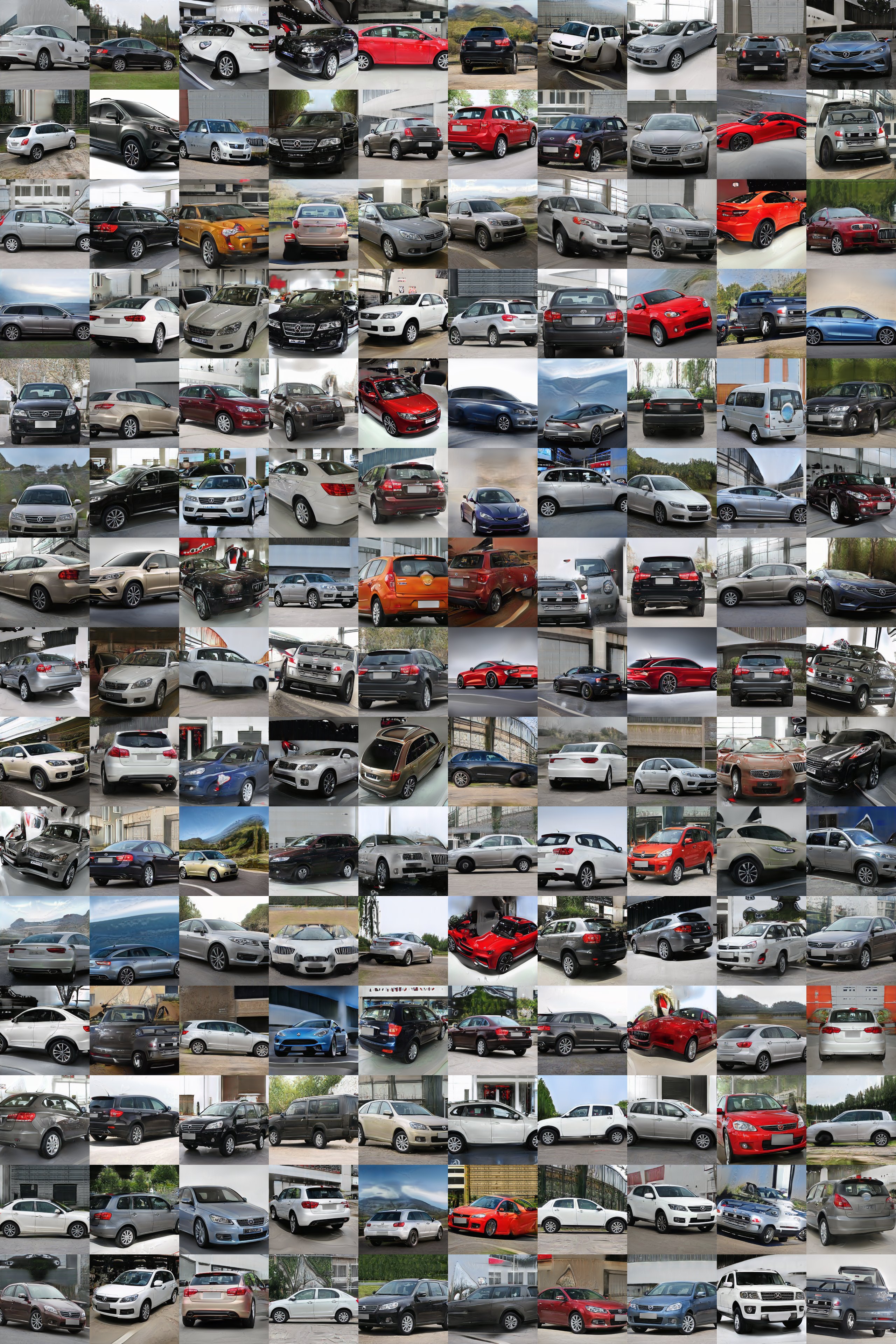}
    \caption{Random samples with random camera poses on CompCars $256^2$.}
    \label{fig:random_compars}
\end{figure}

\end{document}